\documentclass[sigconf]{acmart}
\setcopyright{none}
\renewcommand\footnotetextcopyrightpermission[1]{}
\acmConference{}{}{}
\usepackage{algorithm}
\usepackage{algorithmic}

\usepackage{booktabs}
\usepackage{tabularx}
\usepackage{array}
\usepackage{pifont}

\AtBeginDocument{%
  }

\begin{document}

\title{AgentGFM: A Graph Foundation Model with Node-Agent Information-Flow Control}

\author{Jingbo Cui}

\affiliation{
  \institution{School of Computer Science and Technology, Tianjin University}
  \city{Tianjin}
  \country{China}
}
\email{cjb2025244179@tju.edu.cn}

\author{Jitao Zhao}
\affiliation{
  \institution{School of Computer Science and Technology, Tianjin University}
  \city{Tianjin}
  \country{China}}
\email{zjtao@tju.edu.cn}

\author{Di Jin}
\affiliation{%
  \institution{School of Computer Science and Technology, Tianjin University}
  \city{Tianjin}
  \country{China}}
\email{jindi@tju.edu.cn}

\author{Dongxiao He}
\affiliation{%
  \institution{School of Computer Science and Technology, Tianjin University}
  \city{Tianjin}
  \country{China}}
\email{hedongxiao@tju.edu.cn}

\renewcommand{\shortauthors}{Cui et al.}

\begin{abstract}
Graph Foundation Models (GFMs) aim to learn transferable 
knowledge from multi-domain graphs and adapt to unseen scenarios.
As a fundamental source of relational semantics in graphs, the transferability of topological patterns has long been central to GFM research. However,  local structural patterns may vary across graphs and even among nodes within the same graph.   Despite such structural variation, most existing GFMs rely on manually  designed propagation schemes and apply them to new graphs largely unchanged. Such fixed schemes may not suit the diverse structural patterns of different nodes. This raises a key question: can each node autonomously determine how information should be propagated through the graph? We refer to this capability as information-flow control. Inspired by recent advances in agent technology, we formulate this
problem as agent-based decision making and treat each node as an
agent.  Accordingly, we propose AgentGFM, in which all node agents follow a shared end-to-end trainable policy rather than using independent models. For adaptive information-flow control, each node interacts with the graph through a predict–act–observe–correct process. During the act stage, the node makes three decisions: source reception, signal-channel selection and gain-aware node-wise halting. The resulting observation is compared with the prediction and their discrepancy is used to correct the node state and guide subsequent interactions. Extensive experiments across node-level, graph-level and large-scale
transfer scenarios demonstrate the effectiveness of AgentGFM across
diverse graph topologies.
\end{abstract}



\maketitle

\section{Introduction}

Graph-structured data are ubiquitous in real-world applications, including social network analysis~\cite{appFinance}, recommendation systems~\cite{IF_App_recomend}, protein interaction modeling~\cite{appBio} and knowledge graph reasoning~\cite{r-gcn}.  However, current graph learning models often remain  task-specific, which limits their generality across domains.  Inspired by the success of foundation models in natural language processing and computer vision, recent studies have begun to explore Graph Foundation Models (GFMs). These models aim to learn generalizable graph knowledge from multi-domain graphs, enabling rapid adaptation to new scenarios~\cite{liu2023towards}.

As a fundamental source of relational semantics in graphs, the
transferability of topological patterns has long been central to GFM
research. However, this transferability is challenged by the fact that
local structural patterns may vary across graphs and even among nodes
within the same graph. Existing studies have explored topology-aware designs to improve
generalization under such structural diversity. Prompt-based methods
use topology-aware prompts or structural encodings to adapt pretrained
models to graph-specific
contexts~\cite{GPF/GPF-plus,GraphPrompt,GraphLoRA,SAMGPT}. Other methods
improve transferability through structural routing, topology alignment
or adaptive aggregation~\cite{GraphAny,GFT}. Recent studies further
use Riemannian representations or mixture-of-experts architectures to
model heterogeneous topologies~\cite{RiemannGFM,KDEM,rgfm}.
Transformer-based GFMs also encode graph structures as tokens or graph
sequences for general-purpose modeling~\cite{opengraph,graphgpt}. These efforts have advanced
GFMs toward topology-aware generalization.

\begin{figure}[t]
\centering
\includegraphics[width=\columnwidth]{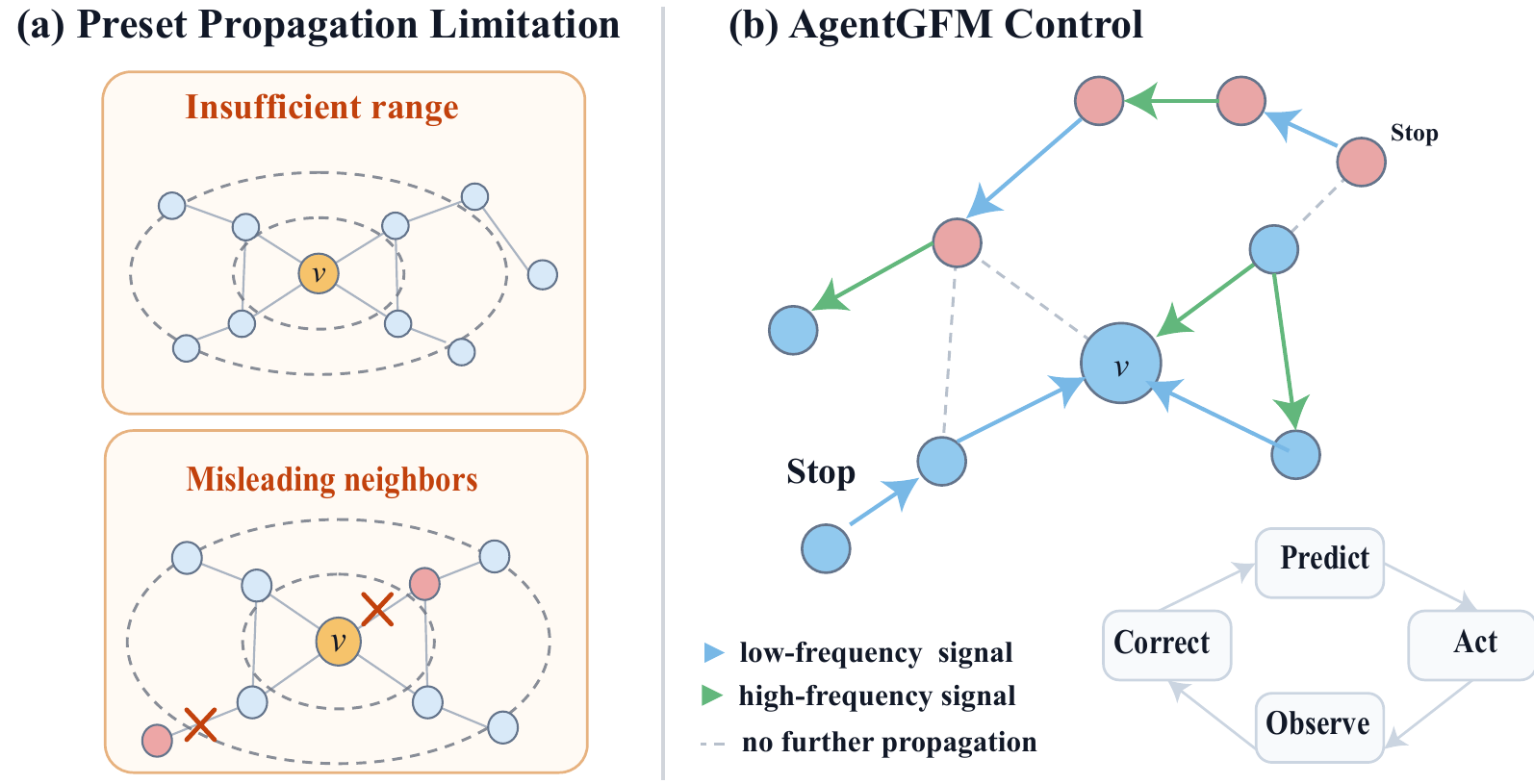}
\Description{
The figure contains two side-by-side diagrams. The left diagram
illustrates preset graph propagation, where a target node may fail to
reach useful distant information or may aggregate misleading neighboring
signals. The right diagram illustrates AgentGFM, where a node performs
a predict--act--observe--correct interaction and selectively controls
information sources, signal channels and halting decisions before
updating its state through feedback.
}
\caption{Motivation of AgentGFM. Preset propagation may miss useful
evidence or absorb noisy signals under topology shifts. AgentGFM instead enables node-level information-flow control through
source reception, signal-channel selection, gain-aware node-wise
halting and prediction--observation feedback.}
\label{fig:motivation}
\end{figure}

Despite this progress, most existing GFMs rely on manually designed propagation schemes and apply them to new graphs largely unchanged. Such fixed schemes may not adapt to the diverse local structural patterns of different nodes. As illustrated in Figure~\ref{fig:motivation}(a), a fixed scheme may stop before useful information is reached or continue to aggregate irrelevant neighboring signals. This mismatch between uniform propagation and node-specific structural patterns limits adaptation under topology shift. This raises a key question: \textbf{can each node autonomously determine how information should be propagated through the graph?} We refer to this capability as \emph{information-flow control}.

However, realizing information-flow control is nontrivial. Propagation decisions must adapt to node-specific structures while remaining transferable across unseen graphs. In addition, expanding the propagation range may capture useful structural dependencies but can also introduce irrelevant signals, making it difficult to determine an appropriate propagation process for each node.
Prior adaptive propagation studies have shown that graph propagation
can be conditioned on graph signals rather than following a uniform
routine~\cite{gdpnet,policygnn}. However, these methods are typically optimized for a specific graph distribution or supervision objective. They are not designed to learn transferable decision mechanisms across graphs.  Existing GFMs adapt to target graphs mainly through
topology-aware prompts or expert routing, while their propagation
schemes remain largely fixed. Thus, how to enable node-specific propagation
decisions that transfer across unseen graphs remains underexplored.

Inspired by recent advances in agent technology, we connect this
problem with agent-based decision making, where agents adapt their
actions according to internal states and environmental feedback. This
paradigm naturally aligns with graph information-flow control, where
each node should determine how information is received and propagated.
Motivated by this perspective, we treat each node as an agent that
makes propagation decisions according to its current state and
refines them through feedback from the graph. Importantly, node agents
do not use independent models. Instead, all nodes follow a shared
end-to-end trainable policy, enabling node-specific decisions while
preserving cross-graph transferability.

Accordingly, we propose AgentGFM, a Graph Foundation Model with
node-agent information-flow control. As illustrated in
Figure~\ref{fig:motivation}(b), each node interacts with the graph
through a predict--act--observe--correct process. It first predicts
the contextual information expected from its current state and then
makes three decisions during the act stage: source reception,
signal-channel selection and gain-aware node-wise halting. The graph
returns an observation induced by these decisions and the discrepancy
between the observation and prediction is used to correct the node
state and guide subsequent interactions. Through this process,
different nodes follow distinct information-flow trajectories under
the same transferable policy.

Our contributions are summarized as follows:
\begin{itemize}
    \item We identify the limitation of transferring manually designed
    propagation schemes to new graphs largely unchanged and formulate
    topology generalization as node-agent information-flow control.

  \item We propose AgentGFM, where each node acts as an agent under a
shared trainable policy and controls information flow through a
predict--act--observe--correct process with source reception,
signal-channel selection and gain-aware halting.

    \item Extensive experiments across node-level, graph-level and
    large-scale transfer scenarios demonstrate the effectiveness of
    AgentGFM across diverse graph topologies.
\end{itemize}

\section{Related Work}

\subsection{Graph Foundation Models}

Graph Foundation Models (GFMs) aim to learn transferable graph
knowledge that generalizes across  graphs. Existing
GFMs mainly address cross-graph heterogeneity from two perspectives:
feature alignment and structural adaptation.

For feature heterogeneity, GCOPE~\cite{GCOPE} introduces coordinator
modules for cross-graph interaction and latent-space alignment, MDGPT
~\cite{mdgpt} uses domain-specific tokens to incorporate domain
information, TIG~\cite{TIG} learns transfer-invariant node features 
and BRIDGE~\cite{BRIDGE} combines feature unification with semantic
alignment. These methods mainly improve transferability by learning
compatible feature spaces.

For structural heterogeneity, GraphControl~\cite{GraphControl}
constructs structure-aware positional encodings from spectral
information, GFT~\cite{GFT} learns transferable propagation patterns,
R-GFM~\cite{rgfm} captures geometry-adaptive representations and
GraphAny~\cite{GraphAny} adopts mixture-of-experts to adapt encoders
to different graph structures.

Although these methods improve GFMs through feature alignment,
structural encoding, or component adaptation, they mainly adapt model
representations or components to target graphs. How individual nodes
should adjust their information propagation process under diverse
structural conditions remains largely unexplored. AgentGFM addresses
this limitation by enabling node-agent information-flow control through
a shared transferable policy.

\subsection{Adaptive Graph Propagation and Decision-based Reasoning}

Adaptive graph propagation and decision-based reasoning methods
learn how information should be acquired over graph structures
instead of relying on fixed message-passing rules. Existing studies
mainly adapt information sources or propagation depths.
DeepPath~\cite{deeppath} and MINERVA~\cite{minerva} formulate
knowledge graph reasoning as path navigation, GDPNet~\cite{gdpnet}
learns adaptive neighborhood selection and
Policy-GNN~\cite{policygnn} learns node-specific propagation
iterations. These studies show that graph reasoning benefits from
adapting where to collect information and how far to propagate.

However, these methods are mainly optimized for a specific task,
graph distribution or supervision objective and are not designed for
cross-graph transfer. In contrast, AgentGFM learns a shared
information-flow policy from multiple source graphs and transfers it
to unseen target graphs. It jointly controls information sources,
signal channels and propagation duration while using
prediction--observation discrepancy for state correction. A
mechanism-level comparison is provided in
Appendix~\ref{app:mechanism_positioning}.

\section{Preliminaries}

\subsection{Problem Setup}

Let $\mathcal{G}=(\mathcal{V},\mathcal{E},\mathbf{X})$ denote a graph,
where $\mathcal{V}$ is the node set, $\mathcal{E}$ is the edge set and
$\mathbf{X}\in\mathbb{R}^{|\mathcal{V}|\times d}$ is the node feature
matrix. We denote the adjacency matrix by $\mathbf{A}$ and the
neighborhood of node $v$ by $\mathcal{N}(v)$. Each node
$v\in\mathcal{V}$ is associated with a feature vector
$\mathbf{x}_v\in\mathbb{R}^{d}$.

We consider a cross-domain GFM setting in which the model is trained
on a set of source graphs $\mathcal{G}_S$ and evaluated on a disjoint
set of target graphs $\mathcal{G}_T$, where
$\mathcal{G}_S\cap\mathcal{G}_T=\varnothing$. The objective is to learn
transferable graph knowledge from $\mathcal{G}_S$ and generalize to
unseen graphs in $\mathcal{G}_T$ with limited supervision.

\subsection{Information-Flow Control Formulation}

Let $\mathbf h_v^{(k)}$ denote the representation of node $v$ at
propagation step $k$. Conventional message passing first aggregates
neighboring representations:
\begin{equation}
\mathbf{m}_v^{(k)}
=
\operatorname{AGG}^{(k)}
\left(
\left\{
\mathbf{h}_u^{(k)}
\mid
u\in\mathcal{N}(v)
\right\}
\right),
\end{equation}
and then updates the node representation:
\begin{equation}
\mathbf{h}_v^{(k+1)}
=
\phi^{(k)}
\left(
\mathbf{h}_v^{(k)},
\mathbf{m}_v^{(k)}
\right),
\end{equation}
where $\operatorname{AGG}^{(k)}$ and $\phi^{(k)}$ are specified by
the model architecture. Consequently, the same propagation mechanism
is typically applied to different nodes and transferred to new graphs
with limited adaptation.

To support node-specific propagation, we formulate information-flow
control as a policy-learning problem. At step $k$, node $v$ constructs
a state
\begin{equation}
\mathbf s_v^{(k)}
=
f_s
\left(
\mathbf h_v^{(k)},
\mathbf m_v^{(k)}
\right),
\end{equation}
and a shared policy produces a node-specific decision:
\begin{equation}
\mathbf d_v^{(k)}
=
\pi_\theta
\left(
\mathbf s_v^{(k)}
\right).
\end{equation}
The representation is then updated through a decision-conditioned
transition:
\begin{equation}
\mathbf h_v^{(k+1)}
=
f_t
\left(
\mathbf h_v^{(k)},
\mathbf m_v^{(k)},
\mathbf d_v^{(k)}
\right).
\end{equation}
This formulation allows different nodes to execute different
propagation decisions while sharing the same policy parameters.
AgentGFM provides a concrete realization of this formulation in
Section~\ref{sec:method}.

\section{Method}
\label{sec:method}

\subsection{Overview of AgentGFM}

\begin{figure*}[t]
    \centering
    \includegraphics[width=\linewidth]{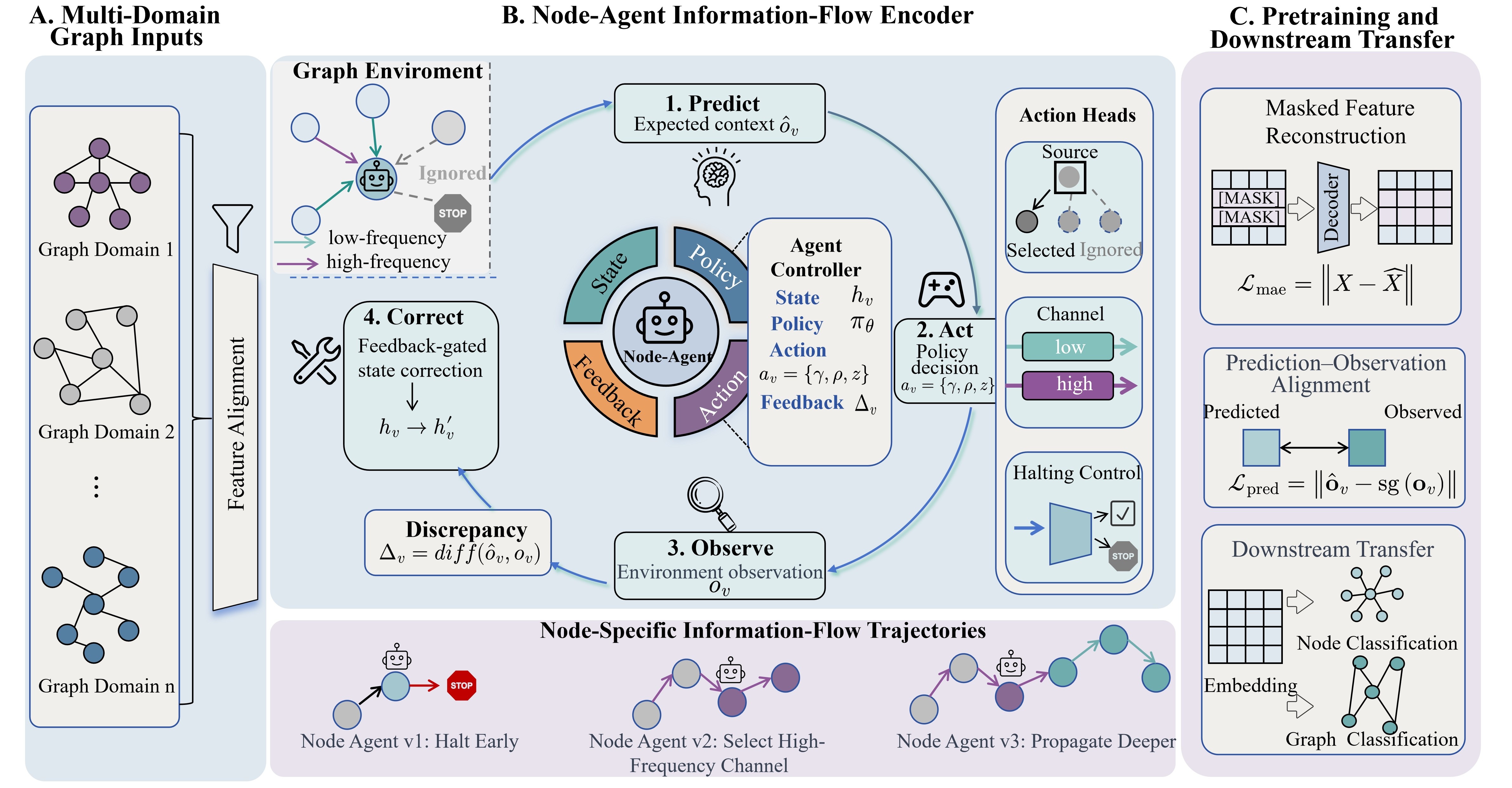}
    \Description{
Overview of AgentGFM. 
Each node acts as an agent that performs predictive observation, information-flow decisions, graph interaction and feedback-based state correction.
}
   \caption{Overall architecture of AgentGFM. Multi-domain node attributes are first aligned into a shared feature space. Each node then performs a predict--act--observe--correct interaction, where information-flow actions control source reception,
signal-channel selection and halting. The encoder is pretrained with masked reconstruction and prediction-observation alignment objectives and transferred to downstream tasks.}
    \label{fig:agentgfm}
\end{figure*}

Given graphs from different domains, AgentGFM first aligns their node
features into a shared space using truncated SVD. The aligned features
are then processed by a shared encoder in which each node acts as an
agent for information-flow control, as illustrated in
Figure~\ref{fig:agentgfm}. The encoder operates through a recurrent
predict--act--observe--correct process. The process is organized into
a small number of outer interaction rounds. Across rounds, each node
updates its state using the feedback from the preceding
prediction--observation comparison, while the act stage performs an
inner information-flow rollout over the graph.
At each round, a node first predicts the information expected from its
current state. It then makes propagation decisions, receives the
resulting information from the graph and updates its state according
to the discrepancy between the prediction and observation. All nodes
follow the same end-to-end trainable policy while making node-specific
decisions. The following subsections introduce the node-level actions,
observation construction, state correction and training objectives.

\subsection{Node-Agent Information-Flow Control}

AgentGFM realizes information-flow control through interactions
between node agents and the graph. Each node acts as an agent, while
the graph serves as the environment that returns contextual
information in response to the node's actions.

A node agent is not an independent model assigned to a node. Instead,
it is a node-specific execution of a shared information-flow policy.
All nodes share the same policy parameters, while their states,
actions, observations and halting decisions remain node-specific.

For node $v_i$, we define its agent state at rollout step $t$ as
\begin{equation}
\mathbf{s}_i^t=\mathbf{c}_i^t,
\end{equation}
where $\mathbf{c}_i^t\in\mathbb{R}^d$ is the carrier state,
$d$ denotes the hidden representation dimension and
$\mathbf{c}_i^0=\mathbf{h}_i$. Given $\mathbf s_i^t$, the shared policy
produces
\begin{equation}
\mathbf d_i^t=
\left(
\boldsymbol{\gamma}_i^t,
\boldsymbol{\rho}_i^t,
z_i^t
\right),
\end{equation}
where $\boldsymbol{\gamma}_i^t$ and $\boldsymbol{\rho}_i^t$ represent
edge-wise source-reception and signal-channel decisions, while $z_i^t$
denotes node-wise halting.

The node agent repeatedly executes these decisions in the graph
environment. At each rollout step, its actions determine how
information is received and processed from neighboring nodes. The
rollout terminates when $z_i^t$ indicates halting or when the maximum
rollout length is reached.

After the rollout, node $v_i$ obtains an observation $\mathbf o_i$
that summarizes the contextual information returned by the graph
environment. This observation is compared with the predicted
observation $\widehat{\mathbf o}_i$ and their discrepancy is used to
correct the node representation:
\begin{equation}
\mathbf h_i'
=
\mathcal T
\left(
\mathbf h_i,
\widehat{\mathbf o}_i,
\mathbf o_i
\right),
\end{equation}
where $\mathcal T(\cdot)$ denotes the state-correction function and
$\mathbf h_i'$ is the corrected representation of node $v_i$.

\subsection{Predictive Observation Estimation}

At the beginning of each interaction round, node $v_i$ predicts the
observation expected from the graph environment based on its current
state. This prediction provides a reference for evaluating the
information returned after the node executes its actions.

Given the current state
$\mathbf{s}_i=\mathbf{h}_i$, the predicted observation is computed
as
\begin{equation}
    \widehat{\mathbf{o}}_i
    =
    f_{\mathrm{pred}}
    \left(
    \mathbf{h}_i
    \right),
    \label{eq:predicted-observation}
\end{equation}
where $f_{\mathrm{pred}}:\mathbb{R}^d\rightarrow\mathbb{R}^d$
is a trainable prediction function. $\widehat{\mathbf{o}}_i$ denotes the observation that node $v_i$
expects to receive from the graph environment.

After the node acts, the graph environment returns an actual
observation $\mathbf{o}_i$. AgentGFM compares
$\widehat{\mathbf{o}}_i$ with $\mathbf{o}_i$ to measure their
discrepancy. This discrepancy reflects the mismatch between the node's
current expectation and the information returned by the graph. It is
used as feedback to correct the node state and guide the propagation
decision in the next interaction round.

\subsection{Policy-Conditioned Information-Flow Rollout}

After predictive observation estimation, each node interacts with the
graph environment through a policy-conditioned information-flow
rollout over the original topology. At rollout step $t$, node $v_i$
decides which sources to receive information from and which signal
channel to use for message transmission.

Let $\mathbf{c}_i^t$ denote the carrier state of node $v_i$ at step
$t$, with $\mathbf{c}_i^0=\mathbf{h}_i$. Each node maintains a
forwarding budget $b_i^t\in[0,1]$ and an activity variable $a_i^t$.
The forwarding budget controls the information transmitted by the
source node, while $a_i^t$ indicates whether node $v_i$ remains active
during the rollout. The activity variable follows a binary halting
trajectory in the forward pass and is optimized through the
straight-through estimator introduced in
Section~\ref{sec:gain_halting}.

For each original edge $(u,i)\in\mathcal{E}$, AgentGFM computes a
source-reception score:
\begin{equation}
\gamma_{u\to i}^t
=
\sigma
\left(
g_\gamma
\left(
\mathbf{c}_u^t,
\mathbf{c}_i^t,
\epsilon_u,
\epsilon_i,
b_u^t
\right)
\right),
\end{equation}
where $\epsilon_u$ and $\epsilon_i$ denote feedback cues from the
preceding prediction--observation interaction. The score
$\gamma_{u\to i}^t$ determines how strongly node $v_i$ receives
information from source node $v_u$ under the current rollout state.
The effective transmission strength along edge $(u,i)$ is defined as
\begin{equation}
\mu_{u\to i}^t
=
a_i^t b_u^t\gamma_{u\to i}^t.
\end{equation}
Accordingly, $\mu_{u\rightarrow i}^t\in[0,1]$ is a scalar effective
transmission strength.
Thus, an active target node receives information according to the
source-reception decision, while a source node transmits information
according to its forwarding budget.

To capture complementary structural signals, AgentGFM constructs
low- and high-frequency channels. Let
$\gamma_{u\to i}^{t-1}$ denote the source-reception score from the
preceding rollout step. For $t>0$, these scores are normalized over
the incoming neighbors:
\begin{equation}
\alpha_{u\to i}^t
=
\frac{
\mathrm{sg}
\left(
\gamma_{u\to i}^{t-1}
\right)
}{
\sum_{v\in\mathcal{N}(i)}
\mathrm{sg}
\left(
\gamma_{v\to i}^{t-1}
\right)
+\varepsilon
},
\end{equation}
where $\mathrm{sg}(\cdot)$ denotes stop-gradient and $\varepsilon>0$ is a small constant for numerical stability.  At the first rollout
step, the weights are initialized uniformly over neighboring nodes.

Using these policy-conditioned weights, the local low- and
high-frequency components are computed as
\begin{equation}
\mathbf{c}_{i,\mathrm{low}}^t
=
\sum_{u\in\mathcal{N}(i)}
\alpha_{u\to i}^t
\mathbf{c}_u^t,
\quad
\mathbf{c}_{i,\mathrm{high}}^t
=
\mathbf{c}_i^t-\mathbf{c}_{i,\mathrm{low}}^t.
\end{equation}
The low-frequency component summarizes contextual information from the
selected sources, while the high-frequency component preserves the
deviation of the node state from its local context.

For source node $v_u$, AgentGFM constructs two candidate messages:
\begin{equation}
\mathbf{m}_{u,\mathrm{low}}^t
=
\mathbf{W}_{\mathrm{low}}
\mathbf{c}_{u,\mathrm{low}}^t,
\quad
\mathbf{m}_{u,\mathrm{high}}^t
=
\mathbf{W}_{\mathrm{high}}
\mathbf{c}_{u,\mathrm{high}}^t.
\end{equation}

The edge-specific channel gate $\rho_{u\to i}^{t}$ is computed from
source--target compatibility, feedback cues and the source forwarding
budget and is mapped into $(0,1)$ through a sigmoid function. It
combines the low- and high-frequency messages as
\begin{equation}
\mathbf{m}_{u\to i}^t
=
\rho_{u\to i}^t
\mathbf{m}_{u,\mathrm{low}}^t
+
\left(
1-\rho_{u\to i}^t
\right)
\lambda_{\mathrm{high}}
\mathbf{m}_{u,\mathrm{high}}^t,
\end{equation}
where $\lambda_{\mathrm{high}}$ controls the contribution of the
high-frequency channel.

Finally, the contextual information received by node $v_i$ at rollout
step $t$ is
\begin{equation}
    \mathbf{r}_i^t
    =
    \sum_{u\in\mathcal{N}(i)}
    \mu_{u\rightarrow i}^t
    \mathbf{m}_{u\rightarrow i}^t
    \in\mathbb{R}^d.
\end{equation}

\subsection{Gain-Aware Halting and Observation Construction}
\label{sec:gain_halting}
After receiving the contextual signal $\mathbf{r}_i^t$, node $v_i$
constructs a candidate carrier state:
\begin{equation}
\widetilde{\mathbf{c}}_i^{t+1}
=
\operatorname{Norm}_2
\left(
\mathbf{c}_i^t
+
f_c
\left(
\left[
\mathbf{c}_i^t
\Vert
\mathbf{r}_i^t
\right]
\right)
\right)
\in\mathbb{R}^d.
\label{eq:candidate_carrier}
\end{equation}

To estimate whether the current information-flow step provides useful
additional information, we introduce a local gain predictor
$f_{\mathrm{gain}}(\cdot)$. It measures the discrepancy between the
received context and the predicted context before and after state
refinement:
\begin{equation}
\begin{aligned}
d_{i,-}^t
&=
\frac{1}{d}
\left\|
\mathbf{r}_i^t
-
f_{\mathrm{gain}}
\left(
\mathbf{c}_i^t
\right)
\right\|_1,\\
d_{i,+}^t
&=
\frac{1}{d}
\left\|
\mathbf{r}_i^t
-
f_{\mathrm{gain}}
\left(
\widetilde{\mathbf{c}}_i^{t+1}
\right)
\right\|_1,\\
g_i^t
&=
d_{i,-}^t-d_{i,+}^t.
\end{aligned}
\label{eq:predictive_gain}
\end{equation}

The resulting predictive gain $g_i^t\in\mathbb{R}$ is node-specific.
A positive gain indicates that the candidate update better captures
the received context, while a small or negative gain suggests limited
benefit from further information acquisition.

Based on the current carrier state, received context, predictive gain,
forwarding budget and rollout-step embedding, a halting policy
generates the halting score:
\begin{equation}
\kappa_i^t
=
f_{\mathrm{halt}}
\left(
\mathbf{c}_i^t,
\mathbf{r}_i^t,
\widetilde{\mathbf{c}}_i^{t+1},
g_i^t,
b_i^t,
\mathbf{e}_t
\right),
\label{eq:halting_score}
\end{equation}
where $\mathbf{e}_t\in\mathbb{R}^{4}$ denotes a learnable embedding
of the current rollout step and $\kappa_i^t\in(0,1)$ is the scalar
halting score.

To retain discrete node-wise execution while enabling end-to-end
optimization, we apply a straight-through estimator:
\begin{equation}
\begin{aligned}
z_{i,\mathrm{h}}^t
&=
\mathbb{I}
\left(
\kappa_i^t\geq\theta_0
\right),\\
z_i^t
&=
z_{i,\mathrm{h}}^t
+
\kappa_i^t
-
\operatorname{sg}
\left(
\kappa_i^t
\right),\\
a_i^{t+1}
&=
a_i^t
\left(
1-z_i^t
\right).
\end{aligned}
\label{eq:straight_through_halting}
\end{equation}

The carrier state is updated according to the activity status at the
current rollout step:
\begin{equation}
\mathbf{c}_i^{t+1}
=
a_i^t
\widetilde{\mathbf{c}}_i^{t+1}
+
\left(
1-a_i^t
\right)
\mathbf{c}_i^t.
\label{eq:carrier_update}
\end{equation}

The forward pass follows the binary halting decision, while gradients
are propagated through the continuous halting score during
optimization. An active node incorporates the contextual information
received at the current rollout step. Once a node halts, its carrier
state remains unchanged in subsequent rollout steps, while the frozen
carrier can still provide source information for active nodes.

Finally, the observation of node $v_i$ is constructed by aggregating
the contextual information collected throughout its node-specific
information-flow rollout:
\begin{equation}
\mathbf{o}_i
=
\operatorname{ReLU}
\left(
\operatorname{LN}
\left(
\sum_{t=0}^{T-1}
\mathbf{r}_i^t
\right)
\right).
\label{eq:observation_construction}
\end{equation}

\subsection{Feedback-Gated State Correction}

After obtaining the actual observation from the graph environment, each node compares it with the observation predicted from its current state. Their discrepancy reflects how far the returned information deviates from the node's expectation. AgentGFM uses this discrepancy to regulate the contribution of the actual observation during state correction.

We define the prediction--observation discrepancy as
\begin{equation}
    \boldsymbol{\delta}_i
    =
    \left|
    \mathbf{o}_i-\widehat{\mathbf{o}}_i
    \right|
    \in\mathbb{R}^{d},
\end{equation}
where $\widehat{\mathbf{o}}_i$ is the predicted observation and
$\mathbf{o}_i$ is the actual observation returned by the graph
environment. This discrepancy is used for the current state correction
and can also provide feedback for subsequent interactions. Based on
$\boldsymbol{\delta}_i$, AgentGFM computes a correction gate:
\begin{equation}
    \mathbf{q}_i
    =
    \sigma\left(
    f_{\mathrm{rel}}
    (
    \mathbf{h}_i,
    \widehat{\mathbf{o}}_i,
    \mathbf{o}_i,
    \boldsymbol{\delta}_i
    )
    \right)
    \in(0,1)^d,
\end{equation}
where $\mathbf{q}_i$ is a feature-wise correction gate that controls
the contribution of each dimension of the actual graph observation.

The corrected observation is obtained through gated fusion:
\begin{equation}
    \widetilde{\mathbf{o}}_i
    =
    \mathbf{q}_i \odot \mathbf{o}_i
    +
    \left(
    1-\mathbf{q}_i
    \right)
    \odot
    \widehat{\mathbf{o}}_i .
    \label{eq:corrected-observation}
\end{equation}
The node state is then updated through a residual correction:
\begin{equation}
\mathbf{h}'_i =
\operatorname{LN}
\left(
\mathbf{h}_i +
\operatorname{Dropout}(\widetilde{\mathbf{o}}_i)
\right),
\label{eq:state_correction}
\end{equation}
where $\mathrm{LN}(\cdot)$ denotes layer normalization.

This feedback-gated correction allows the node to balance the actual
graph observation with its prediction before updating the state.
Unlike directly weighting messages by their transmission strength, the
correction gate is conditioned on the discrepancy between predicted and
observed information, making the state update explicitly
feedback-driven.

\begin{table*}[t]
\centering
\caption{Cross-domain 1-shot node classification accuracy (\%). The
best result on each dataset is highlighted in bold. Avg. Rank denotes
the average ranking across all datasets, where a lower value is better.}
\Description{
A comparison of fourteen graph learning methods across ten node
classification datasets under the cross-domain 1-shot protocol.
Methods are grouped into task-supervised GNNs, self-supervised
pretraining methods and graph foundation models. AgentGFM achieves
the best average rank and the highest accuracy on nine datasets.
}
\label{tab:cross_domain_1shot}

\resizebox{\textwidth}{!}{
\begin{tabular}{@{}lccccccccccc@{}}
\toprule
\textbf{Method}
& \textbf{Cora}
& \textbf{CiteSeer}
& \textbf{PubMed}
& \textbf{Computers}
& \textbf{Photo}
& \textbf{Texas}
& \textbf{Wisconsin}
& \textbf{Cornell}
& \textbf{Chameleon}
& \textbf{Squirrel}
& \textbf{Avg. Rank} \\
\midrule

\multicolumn{12}{@{}l}{\textbf{Task-Supervised GNNs}} \\
\midrule
GCN
& $35.08 \pm 8.90$
& $27.96 \pm 7.77$
& $49.15 \pm 8.93$
& $26.96 \pm 7.92$
& $38.63 \pm 7.60$
& $34.39 \pm 15.53$
& $26.90 \pm 9.89$
& $26.96 \pm 8.89$
& $21.26 \pm 2.57$
& $20.71 \pm 1.57$
& $8.65$ \\

GAT
& $35.90 \pm 10.40$
& $28.85 \pm 8.79$
& $47.55 \pm 11.05$
& $28.52 \pm 14.48$
& $34.10 \pm 9.63$
& $32.85 \pm 15.65$
& $25.82 \pm 10.89$
& $26.58 \pm 10.58$
& $21.02 \pm 2.52$
& $20.26 \pm 0.94$
& $9.90$ \\

FAGCN
& $36.93 \pm 9.00$
& $28.05 \pm 8.79$
& $50.82 \pm 8.64$
& $12.47 \pm 5.92$
& $23.33 \pm 7.72$
& $32.65 \pm 15.90$
& $27.64 \pm 10.16$
& $26.96 \pm 8.83$
& $21.15 \pm 2.89$
& $20.57 \pm 2.40$
& $8.85$ \\

GPRGNN
& $41.31 \pm 10.45$
& $30.86 \pm 10.10$
& $49.17 \pm 10.89$
& $33.57 \pm 15.22$
& $55.09 \pm 13.80$
& $32.75 \pm 14.71$
& $24.50 \pm 8.71$
& $25.78 \pm 7.27$
& $21.10 \pm 2.70$
& $20.17 \pm 0.92$
& $8.80$ \\

\midrule
\multicolumn{12}{@{}l}{\textbf{Self-Supervised Pretraining}} \\
\midrule
DGI
& $24.47 \pm 4.58$
& $25.80 \pm 4.11$
& $40.60 \pm 5.42$
& $35.36 \pm 9.76$
& $47.97 \pm 7.02$
& $18.49 \pm 8.87$
& $20.33 \pm 11.03$
& $16.53 \pm 8.06$
& $21.43 \pm 2.28$
& $20.44 \pm 1.25$
& $10.80$ \\

GraphCL
& $29.52 \pm 5.26$
& $26.68 \pm 4.13$
& $40.19 \pm 7.29$
& $37.74 \pm 8.27$
& $47.85 \pm 7.58$
& $20.14 \pm 12.97$
& $28.39 \pm 10.38$
& $18.38 \pm 8.16$
& $22.89 \pm 3.26$
& $21.16 \pm 1.84$
& $9.20$ \\

GraphMAE
& $41.82 \pm 6.57$
& $38.32 \pm 9.03$
& $48.68 \pm 8.29$
& $43.51 \pm 9.68$
& $60.25 \pm 9.01$
& $36.16 \pm 15.72$
& $25.38 \pm 8.62$
& $22.84 \pm 7.37$
& $21.35 \pm 3.41$
& $21.05 \pm 3.99$
& $6.30$ \\

\midrule
\multicolumn{12}{@{}l}{\textbf{Graph Foundation Models}} \\
\midrule
SAMGPT
& $42.02 \pm 6.01$
& $39.11 \pm 6.92$
& $46.49 \pm 6.97$
& $46.48 \pm 8.67$
& $58.78 \pm 8.62$
& $27.46 \pm 7.78$
& $31.01 \pm 7.78$
& $30.54 \pm 7.07$
& $29.37 \pm 3.73$
& $22.60 \pm 2.82$
& $5.30$ \\

BRIDGE
& $44.43 \pm 6.28$
& $40.05 \pm 6.49$
& $50.52 \pm 7.53$
& $39.67 \pm 7.77$
& $56.44 \pm 8.89$
& $31.53 \pm 7.96$
& $34.26 \pm 8.27$
& $34.33 \pm 7.41$
& $28.92 \pm 4.59$
& $22.52 \pm 3.56$
& $4.40$ \\

R-GFM
& $42.97 \pm 5.41$
& $\mathbf{51.25 \pm 12.41}$
& $48.38 \pm 9.25$
& $25.44 \pm 5.66$
& $57.20 \pm 7.22$
& $32.98 \pm 7.00$
& $26.98 \pm 5.58$
& $21.12 \pm 6.04$
& $21.83 \pm 1.93$
& $20.29 \pm 0.80$
& $7.20$ \\

GCOPE
& $43.33 \pm 10.09$
& $36.61 \pm 6.61$
& $39.45 \pm 8.18$
& $28.94 \pm 7.98$
& $40.86 \pm 6.27$
& $30.01 \pm 8.33$
& $36.21 \pm 12.38$
& $32.70 \pm 7.43$
& $30.32 \pm 4.56$
& $23.77 \pm 3.62$
& $6.20$ \\

GraphAny
& $39.25 \pm 7.18$
& $35.77 \pm 6.76$
& $52.16 \pm 8.33$
& $49.11 \pm 12.02$
& $61.51 \pm 9.95$
& $34.74 \pm 17.75$
& $41.37 \pm 13.65$
& $34.59 \pm 12.46$
& $25.40 \pm 4.12$
& $20.42 \pm 1.63$
& $4.30$ \\

AgentGFM
& $\mathbf{51.94 \pm 7.28}$
& $44.32 \pm 8.20$
& $\mathbf{52.96 \pm 9.15}$
& $\mathbf{55.93 \pm 9.52}$
& $\mathbf{67.48 \pm 8.64}$
& $\mathbf{42.21 \pm 13.20}$
& $\mathbf{44.57 \pm 12.95}$
& $\mathbf{39.83 \pm 9.40}$
& $\mathbf{30.54 \pm 4.17}$
& $\mathbf{24.20 \pm 3.50}$
& $\mathbf{1.10}$ \\

\bottomrule
\end{tabular}
}
\end{table*}

\subsection{Training Objective and Optimization}
\label{sec:objective}

AgentGFM is pretrained on source graphs using self-supervised
objectives. Given a masked node set $\mathcal M$, the masked attribute
reconstruction objective is
\begin{equation}
\mathcal L_{\mathrm{mae}}
=
\frac{1}{|\mathcal M|}
\sum_{v_i\in\mathcal M}
\left\|
\mathbf x_i-
f_{\mathrm{dec}}(\mathbf h_i')
\right\|_1 .
\label{eq:mae_loss}
\end{equation}
This objective encourages the model to recover masked attributes using
contextual information from the graph rather than relying only on
visible node features.

To train predictive observation estimation, we align the predicted
observation with the actual observation returned by the graph
environment:
\begin{equation}
\mathcal L_{\mathrm{pred}}
=
\frac{1}{|\mathcal V|}
\sum_{v_i\in\mathcal V}
\left(
1-
\frac{
\widehat{\mathbf o}_i^\top
\operatorname{sg}(\mathbf o_i)
}{
\|\widehat{\mathbf o}_i\|_2
\|\mathbf o_i\|_2+\varepsilon
}
\right),
\label{eq:prediction_loss}
\end{equation}
where $\operatorname{sg}(\cdot)$ denotes stop-gradient.

To prevent excessive information propagation, we regularize source
reception, effective transmission strength and rollout length:
\begin{equation}
\mathcal L_{\gamma}
=
\mathbb E_{t,(u,i)\in\mathcal E}
\left[\gamma_{u\rightarrow i}^{t}\right],
\quad
\mathcal L_{\mu}
=
\mathbb E_{t,(u,i)\in\mathcal E}
\left[\mu_{u\rightarrow i}^{t}\right],
\quad
\mathcal L_a
=
\mathbb E_{t,i}
\left[a_i^{t+1}\right].
\label{eq:policy_regularizers}
\end{equation}
These terms discourage uniformly strong source reception, excessive
information transmission and unnecessarily long rollouts,
respectively.

The final training objective is
\begin{equation}
\mathcal L
=
\lambda_{\mathrm{mae}}\mathcal L_{\mathrm{mae}}
+
\lambda_{\mathrm{pred}}\mathcal L_{\mathrm{pred}}
+
\lambda_{\gamma}\mathcal L_{\gamma}
+
\lambda_{\mu}\mathcal L_{\mu}
+
\lambda_a\mathcal L_a .
\label{eq:total_objective}
\end{equation}

\paragraph{Implementation of learnable components.} 
Observation prediction, carrier updating, gain estimation, gain-aware halting and reliability gating are implemented using lightweight two-layer MLPs, while the edge-level source-reception and low-/high-frequency channel gates combine source–target query–key projections with scalar MLPs. The detailed parameterization, normalization operations and parameter-sharing scheme are provided in Appendix~\ref{app:information_flow_policy}.

\section{Experiments}

\subsection{Experimental Setup}

\subsubsection{Datasets and Tasks}
We evaluate AgentGFM on the cross-domain 1-shot setting. For node classification, we use ten  datasets, including Texas, Wisconsin, Cornell, Chameleon, Squirrel~\cite{Geom-GCN}, Cora, CiteSeer, PubMed~\cite{citation_cora_citeseer_pubmed}, Computers and Photo~\cite{amazon_computers_photo-coauthor_cs-physics}. These datasets cover diverse domains and topological patterns, including both homophilous and heterophilous graphs. For graph classification, we evaluate on MUTAG, DD~\cite{MUTAG_DD}, IMDB-BINARY~\cite{COLLAB_IMDB}, ENZYMES~\cite{ENZYMES} and PROTEINS~\cite{Protein}. To further examine scalability, we conduct large-scale node classification experiments on Physics~\cite{amazon_computers_photo-coauthor_cs-physics}, Ogbn-Products and Ogbn-Arxiv~\cite{dataOGB}.

\subsubsection{Evaluation Protocol}

We adopt a leave-one-dataset-out cross-domain 1-shot protocol. For each
target dataset, the model is pretrained only on the remaining source
datasets, ensuring that the target graph is unseen during pretraining.
For downstream evaluation, we randomly sample one labeled instance per
class as the support set and use its representation as the class
prototype. Each remaining instance is assigned to the prototype with
the highest cosine similarity. We independently sample the support set
100 times and report the mean accuracy and standard deviation. For graph classification, graph representations are obtained by mean
pooling node representations and one labeled graph per class is used
to construct the prototypes. All methods are evaluated under identical source--target splits and 1-shot settings.

\subsubsection{Baselines}
We compare AgentGFM with three groups of methods. The first group includes task-supervised GNNs, such as GCN~\cite{GCN}, GAT~\cite{GAT}, FAGCN~\cite{fagcn}  and GPRGNN~\cite{GPR-GNN}. The second group includes self-supervised graph pretraining methods, such as DGI~\cite{DGI}, GraphCL~\cite{GRAPHCL} and GraphMAE~\cite{GraphMAE}. The third group consists of GFMs, including SAMGPT~\cite{SAMGPT}, BRIDGE~\cite{BRIDGE}, R-GFM~\cite{rgfm}, GCOPE~\cite{GCOPE} and GraphAny~\cite{GraphAny}. All methods are evaluated under the same  protocol.

\subsection{Performance Analysis}

\subsubsection{Cross-Domain 1-Shot Node Classification}
Table~\ref{tab:cross_domain_1shot} reports the cross-domain 1-shot node classification results. AgentGFM achieves the best average rank of 1.10 across ten target datasets. It obtains the best performance on nine datasets and remains competitive on CiteSeer. The improvement is especially clear on heterophilous graphs, where fixed propagation rules
are more likely to absorb misleading neighboring signals. Baselines exhibit unstable rankings across target datasets, especially under heterophilous settings. In contrast, AgentGFM remains consistently competitive across both homophilous and heterophilous graphs, demonstrating the robustness of topology-adaptive node-agent information-flow control under diverse structural patterns.

\begin{table}[t]
\centering
\caption{Cross-domain 1-shot graph classification accuracy (\%).}
\label{tab:cross_domain_graph_1shot}
\resizebox{\columnwidth}{!}{
\begin{tabular}{@{}lcccccc@{}}
\toprule
\textbf{Method}
& \textbf{MUTAG}
& \textbf{IMDB-B}
& \textbf{ENZYMES}
& \textbf{PROTEINS}
& \textbf{DD}
& \textbf{Avg. Rank} \\
\midrule
DGI
& $58.26{\pm}17.70$
& $51.24{\pm}5.88$
& $18.96{\pm}3.17$
& $54.90{\pm}9.80$
& $53.67{\pm}6.27$
& $5.60$ \\

GraphCL
& $56.65{\pm}15.85$
& $51.64{\pm}6.11$
& $19.13{\pm}3.09$
& $55.93{\pm}8.39$
& $54.85{\pm}8.37$
& $4.60$ \\

GraphMAE
& $56.72{\pm}13.86$
& $50.78{\pm}4.81$
& $18.66{\pm}2.69$
& $54.77{\pm}9.64$
& $53.41{\pm}6.50$
& $7.60$ \\

SAMGPT
& $57.17{\pm}13.91$
& $50.80{\pm}3.55$
& $18.90{\pm}4.00$
& $55.38{\pm}10.87$
& $56.24{\pm}7.00$
& $5.20$ \\

BRIDGE
& $55.88{\pm}14.25$
& $52.37{\pm}5.88$
& $20.77{\pm}3.70$
& $55.55{\pm}10.71$
& $56.02{\pm}8.93$
& $3.80$ \\

R-GFM
& $58.30{\pm}15.00$
& $51.15{\pm}4.40$
& $20.44{\pm}3.19$
& $55.89{\pm}11.25$
& $55.93{\pm}10.98$
& $3.60$ \\

GCOPE
& $57.95{\pm}12.60$
& $51.57{\pm}10.08$
& $19.80{\pm}3.44$
& $55.25{\pm}10.29$
& $55.27{\pm}8.79$
& $4.60$ \\

AgentGFM
& $\mathbf{59.52{\pm}11.99}$
& $\mathbf{52.59{\pm}7.35}$
& $\mathbf{21.50{\pm}2.64}$
& $\mathbf{56.87{\pm}13.01}$
& $\mathbf{56.50{\pm}7.11}$
& $\mathbf{1.00}$ \\
\bottomrule
\end{tabular}
}
\end{table}
\begin{table}[t]
\centering
\caption{Cross-domain 1-shot node classification accuracy (\%) on large-scale datasets. The best result on each dataset is highlighted in bold.}
\Description{Cross-domain one-shot node classification accuracy comparison across three large-scale node classification datasets.}
\label{tab:cross_domain_large_1shot}
\begin{tabular}{@{}lccc@{}}
\toprule
\textbf{Method}
& \textbf{Physics}
& \textbf{Products}
& \textbf{Arxiv} \\
\midrule
DGI
& $50.17{\pm}13.16$
& $9.84{\pm}2.72$
& $13.07{\pm}2.76$ \\

GraphCL
& $55.03{\pm}9.66$
& $12.54{\pm}3.17$
& $17.30{\pm}4.45$ \\

GraphMAE
& $70.78{\pm}13.00$
& $4.17{\pm}1.53$
& $10.32{\pm}2.85$ \\

SAMGPT
& $63.86{\pm}8.55$
& $15.23{\pm}3.34$
& $15.81{\pm}3.26$ \\

BRIDGE
& $69.18{\pm}9.79$
& $16.25{\pm}3.74$
& $16.97{\pm}3.94$ \\

R-GFM
& $54.25{\pm}8.76$
& OOM
& OOM \\

GCOPE
& $58.89 {\pm}9.31$
& $10.65 {\pm} 2.44$
& $12.89 {\pm} 3.42$ \\

GraphAny
& $\mathbf{74.18{\pm}12.92}$
& $10.89{\pm}1.84$
& $17.99{\pm}5.08$ \\

AgentGFM
& $72.36{\pm}7.08$
& $\mathbf{18.21{\pm}3.55}$
& $\mathbf{20.94{\pm}3.33}$ \\
\bottomrule
\end{tabular}
\end{table}

\subsubsection{Cross-Domain 1-Shot Graph Classification}
Table~\ref{tab:cross_domain_graph_1shot} reports the cross-domain 1-shot graph classification results. AgentGFM ranks first on all five datasets and achieves the best average rank of 1.00. Although the margins over the strongest baselines are modest on some datasets, the consistent
improvements across molecular, bioinformatics and social graph benchmarks demonstrate that the transferability of AgentGFM extends to graph-level tasks. We also observe that the strongest baseline varies across datasets, whereas AgentGFM maintains consistently strong performance, suggesting that node-agent information-flow control
provides stable graph-level transfer.

\subsubsection{Cross-Domain 1-Shot Large-Scale Evaluation}
Table~\ref{tab:cross_domain_large_1shot} reports the results on large-scale node classification datasets. AgentGFM is pretrained on eight source graphs, including Cora, CiteSeer, PubMed, Computers, Photo, Texas, Wisconsin and Cornell and then transferred to Physics, Ogbn-Products and Ogbn-Arxiv. AgentGFM achieves the best results on
Ogbn-Products and Ogbn-Arxiv and remains competitive on Physics. Notably, R-GFM runs out of memory on Ogbn-Products and Ogbn-Arxiv,
whereas AgentGFM completes evaluation on both datasets. AgentGFM
performs its rollout over the original edge set without explicitly
constructing dense higher-order neighborhoods, which helps retain
practical scalability on large graphs.
These results demonstrate strong
cross-domain transfer performance on large graphs.

\subsection{Ablation Study}

\begin{figure}[t]
\centering
\includegraphics[width=1\linewidth]{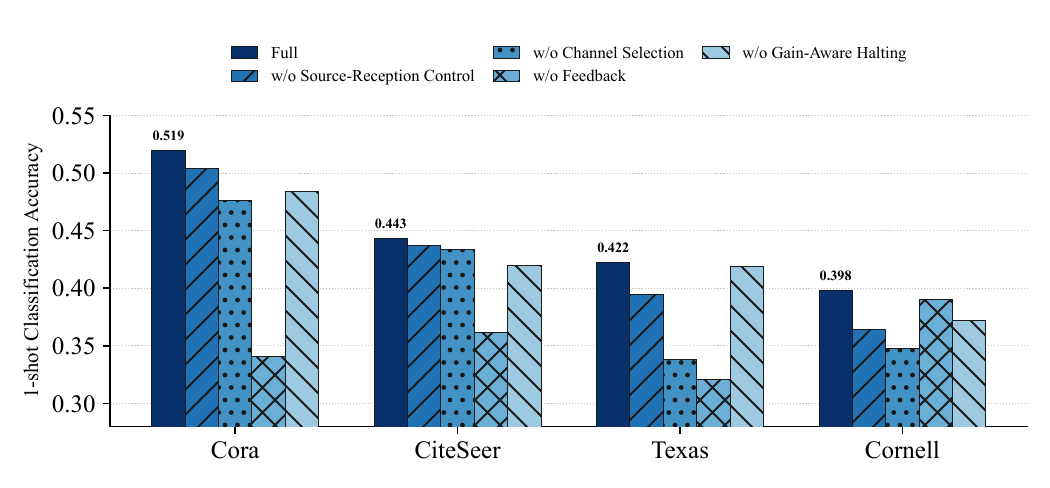}
\Description{
Bar chart showing component-wise ablation results of AgentGFM on representative cross-domain node classification datasets.
The full model is compared with variants removing source reception, signal-channel selection, feedback, or halting control.
}
\caption{Mechanism-aligned ablation study on cross-domain 1-shot node
classification.}
\label{fig:component_ablation}
\end{figure}

\begin{figure}[t]
\centering
\includegraphics[width=0.96\linewidth]{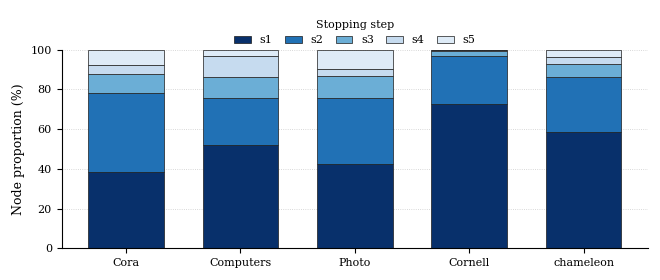}
\Description{
Distribution of node-specific stopping steps under the maximum rollout horizon.
The figure shows the proportion of nodes terminating at different propagation steps across multiple datasets.
}
\caption{Distribution of node-specific halting steps in AgentGFM.}
\label{fig:node_stop_distribution}
\end{figure}

\subsubsection{Mechanism-Aligned Ablation}
Following the mechanism-level comparison in
Table~\ref{tab:propagation_positioning}, we evaluate four ablations
that isolate the key control mechanisms of AgentGFM:
\textit{w/o Source-Reception Control}, \textit{w/o Channel Selection},
\textit{w/o Feedback} and \textit{w/o Gain-Aware Halting}. These
variants remove node-specific source reception, signal-channel
selection, prediction--observation feedback and adaptive node-wise
halting respectively. Figure~\ref{fig:component_ablation} presents
the results on four representative datasets, while the complete results
on all ten node classification datasets are reported in
Appendix~\ref{app:full_ablation}.

As shown in Figure~\ref{fig:component_ablation}, the full model achieves
the best performance on all four datasets. Removing Source-Reception
Control consistently reduces accuracy, supporting the contribution of
node-specific reception. Removing Channel Selection also degrades
performance, with clearer effects on Texas and Cornell, indicating the
benefit of adapting signal channels across different graph structures.
Removing Feedback causes the largest drops on Cora, CiteSeer and Texas,
highlighting the role of prediction--observation feedback in state
correction. Gain-Aware Halting further improves performance by allowing
nodes to use different rollout lengths. These results empirically
support the mechanism-level distinctions summarized in
Table~\ref{tab:propagation_positioning}.

\begin{table}[t]
\centering
\caption{Hyperparameter analysis of the maximum rollout horizon $T_{\max}$. The best result on each dataset is highlighted in bold. Lower average rank indicates better overall performance.}
\label{tab:rollout-steps}
\resizebox{\linewidth}{!}{
\begin{tabular}{lccccc}
\toprule
Dataset & $T_{\max}=2$ & $T_{\max}=3$ & $T_{\max}=4$ & $T_{\max}=5$ & $T_{\max}=6$ \\
\midrule
Cora        & 0.5140 & 0.5153 & 0.5154 & 0.5194 & \textbf{0.5221} \\
CiteSeer    & 0.4136 & 0.4291 & 0.4297 & \textbf{0.4432} & 0.4343 \\
PubMed      & 0.5200 & 0.5182 & 0.5028 & \textbf{0.5296} & 0.5101 \\
Computers   & 0.5504 & 0.5465 & 0.5442 & \textbf{0.5593} & 0.5467 \\
Photo       & 0.6692 & 0.6726 & 0.6627 & 0.6748 & \textbf{0.6858} \\
Texas       & 0.4049 & 0.3924 & 0.4105 & \textbf{0.4221} & 0.4153 \\
Wisconsin   & 0.4339 & 0.4400 & 0.4287 & \textbf{0.4457} & 0.4310 \\
Cornell     & 0.4001 & 0.3967 & \textbf{0.4090} & 0.3983 & 0.3913 \\
Chameleon   & 0.2869 & 0.2891 & 0.2882 & \textbf{0.3054} & 0.3027 \\
Squirrel    & 0.2300 & 0.2330 & 0.2341 & \textbf{0.2420} & 0.2305 \\
\midrule
Avg. Rank   & 3.7 & 3.5 & 3.6 & \textbf{1.4} & 2.8 \\
\bottomrule
\end{tabular}
}
\end{table}

\begin{figure}[t]
    \centering
    \includegraphics[width=\linewidth]{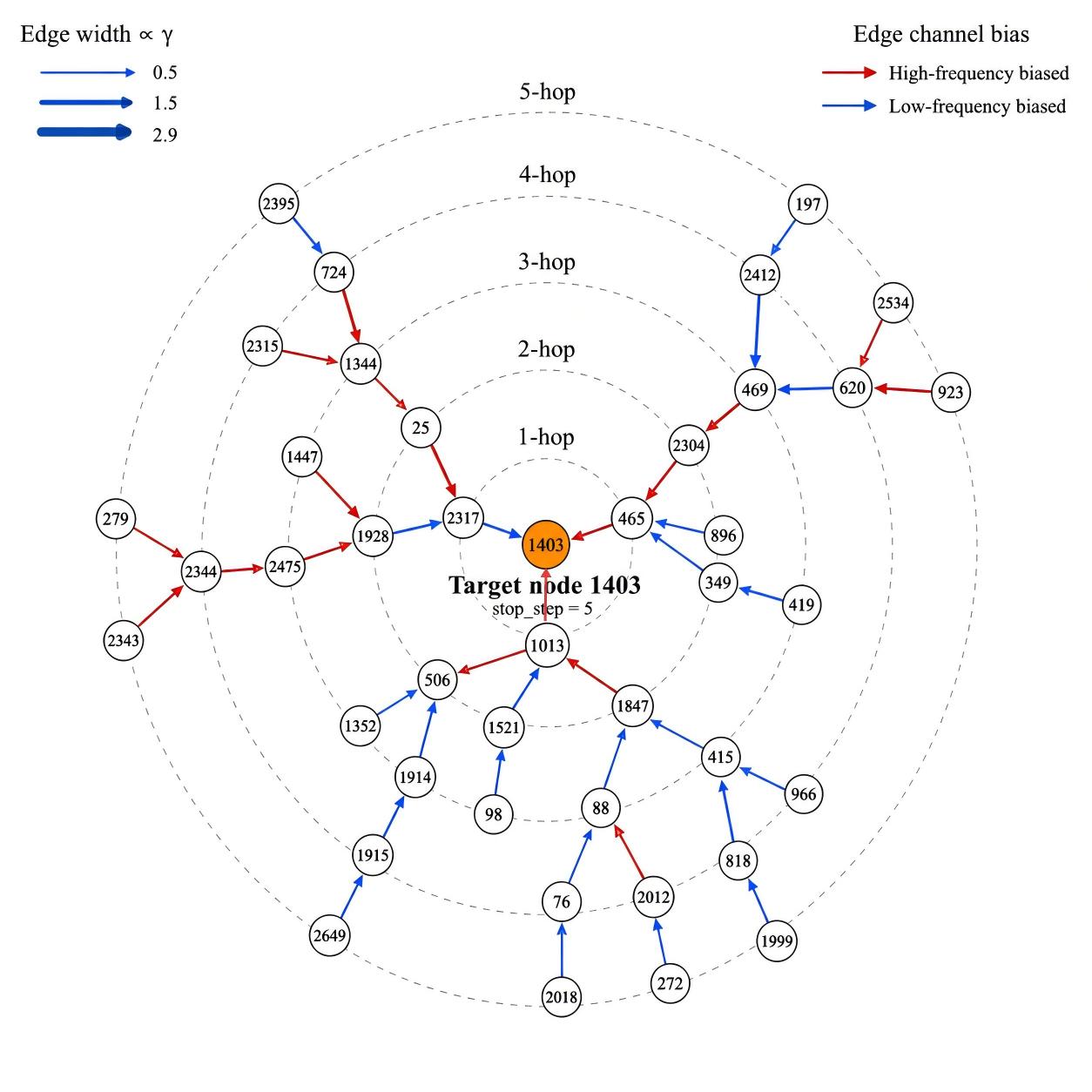}
    \Description{
Visualization of the learned information-flow trajectory for target node 1403 from the Cora dataset.
The figure shows selected source nodes, propagation hops, edge reception strengths and channel preferences.
}
    \caption{
Node-agent information-flow trace for target node 1403 in Cora.
Edge width denotes source-reception strength $\gamma$ and color
indicates channel preference.
    }   
    \label{fig:case_node1403}
\end{figure}

\subsubsection{Node-Specific Information-Flow Depth}
Figure~\ref{fig:node_stop_distribution} shows the distribution of
node-specific halting steps under a maximum rollout horizon of $T=5$.
Most nodes halt before reaching the maximum horizon. On average,
82.42\% of nodes halt by the second rollout step, showing that
AgentGFM does not apply the full rollout to every node.

The distributions also vary across datasets. On Cornell, 96.72\% of
nodes halt by the second step. In contrast, Cora, Computers and Photo
retain a larger proportion of nodes at later steps, with 12.52\%,
13.69\% and 13.13\% of nodes halting at steps 4 or 5 respectively.
These differences show that AgentGFM does not rely on a single global
rollout depth. Instead, different nodes terminate according to their
states and estimated propagation gains. Together with the ablation
results, this analysis confirms that gain-aware halting actively
controls the duration of node-level information acquisition.

\subsubsection{Node-Agent Information-Flow Case Study}
Figure~\ref{fig:case_node1403} visualizes the learned information-flow
trace of target node 1403 in Cora. Each edge represents a
source-reception decision, with edge width indicating reception
strength and color denoting channel preference. AgentGFM selectively
collects information along multiple structural branches, reaching
nodes up to five hops away. The varying path lengths, reception
strengths and channel preferences show that the shared policy produces
a node-specific propagation pattern rather than applying a uniform
aggregation rule.

\subsubsection{Sensitivity to Rollout Horizon}

Table~\ref{tab:rollout-steps} reports the sensitivity of AgentGFM to the maximum rollout horizon $T_{\max}$. Setting $T_{\max}=5$ achieves the
best average rank and the best performance on seven of ten datasets,
showing that a moderate horizon provides sufficient context for most
node agents. Smaller horizons may restrict information acquisition,
whereas increasing $T_{\max}$ to 6 brings no consistent improvement and
may introduce less useful contextual signals. These results support an
adaptive but bounded information-flow rollout and we use
$T_{\max}=5$ as the default setting.

\section{Efficiency and Complexity Analysis}

AgentGFM performs recurrent information-flow rollouts over the
original edge set. Let $R$ denote the number of outer
predict--act--observe--correct interaction rounds, $T$ the maximum
rollout horizon within each round and $d$ the hidden dimension.
Source reception, channel selection and message aggregation are
computed along existing edges at each rollout step, resulting in a
per-step complexity of $O(|E|d)$ and an overall information-flow
complexity of $O(RT|E|d)$. During each rollout step, the node states
and edge-level control variables require $O(|V|d+|E|)$ memory. Since
AgentGFM does not explicitly construct dense higher-order
neighborhoods or all-pair structural contexts, it remains scalable for
large graphs when $R$ and $T$ are bounded by small constants.
\begin{table}[t]
\centering
\caption{Efficiency comparison when pretraining on eight source datasets and evaluating on ogbn-arxiv.}
\label{tab:efficiency}
\begin{tabular}{lcccc}
\toprule
Method & Train Time & Train Mem. & Test Time & Test Mem. \\
\midrule
BRIDGE & 0.72 s & 19.62 GB & 0.35 s & 1.58 GB \\
SAMGPT & 0.53 s & 8.14 GB & 1.24 s & 2.35 GB \\
GraphAny & 1.75 s & 0.67 GB & 18.98 s & 1.62 GB \\
AgentGFM & 1.14 s & 10.21 GB & 0.89 s & 3.24 GB \\
\bottomrule
\end{tabular}
\end{table}
Table~\ref{tab:efficiency} reports the empirical efficiency comparison
when pretraining on eight source datasets and evaluating on
ogbn-arxiv. AgentGFM takes 1.14 seconds per pretraining epoch and
10.21 GB of training memory, which is more memory-efficient than
BRIDGE while remaining practical in training time. During target
evaluation, AgentGFM takes 0.89 seconds with 3.24 GB of memory,
making it faster than GraphAny and SAMGPT at inference. Although
AgentGFM is not the fastest or most memory-efficient method in every
aspect, it achieves a balanced efficiency profile while retaining
adaptive node-agent information-flow control.

\section{Conclusion}

In this paper, we study topology generalization in GFMs through
node-agent information-flow control. Existing GFMs typically transfer
predefined propagation schemes to new graphs with limited
adaptation, which can be restrictive when nodes require different
propagation behaviors under unseen topologies. We therefore propose
AgentGFM, which treats each node as an agent interacting with the graph
environment under a shared transferable policy. Through a
predict--act--observe--correct process, AgentGFM enables node-specific
source reception, signal-channel selection and gain-aware halting,
followed by feedback-gated state correction. Experiments on
cross-domain node classification, graph classification and large-scale
node classification demonstrate consistent transfer performance across
diverse graph structures.

AgentGFM currently incurs additional computation due to its recurrent
information-flow rollout. Future work will investigate more efficient
interaction and halting mechanisms while preserving node-level
adaptability and cross-graph transferability.

\bibliographystyle{ACM-Reference-Format}
\bibliography{references}


\begin{thebibliography}{40}


\ifx \showCODEN    \undefined \def \showCODEN     #1{\unskip}     \fi
\ifx \showISBNx    \undefined \def \showISBNx     #1{\unskip}     \fi
\ifx \showISBNxiii \undefined \def \showISBNxiii  #1{\unskip}     \fi
\ifx \showISSN     \undefined \def \showISSN      #1{\unskip}     \fi
\ifx \showLCCN     \undefined \def \showLCCN      #1{\unskip}     \fi
\ifx \shownote     \undefined \def \shownote      #1{#1}          \fi
\ifx \showarticletitle \undefined \def \showarticletitle #1{#1}   \fi
\ifx \showURL      \undefined \def \showURL       {\relax}        \fi
\providecommand\bibfield[2]{#2}
\providecommand\bibinfo[2]{#2}
\providecommand\natexlab[1]{#1}
\providecommand\showeprint[2][]{arXiv:#2}

\bibitem[Bo et~al\mbox{.}(2021)]%
        {fagcn}
\bibfield{author}{\bibinfo{person}{Deyu Bo}, \bibinfo{person}{Xiao Wang},
  \bibinfo{person}{Chuan Shi}, {and} \bibinfo{person}{Huawei Shen}.}
  \bibinfo{year}{2021}\natexlab{}.
\newblock \showarticletitle{Beyond Low-frequency Information in Graph
  Convolutional Networks}. In \bibinfo{booktitle}{\emph{Thirty-Fifth {AAAI}
  Conference on Artificial Intelligence, {AAAI} 2021, Thirty-Third Conference
  on Innovative Applications of Artificial Intelligence, {IAAI} 2021, The
  Eleventh Symposium on Educational Advances in Artificial Intelligence, {EAAI}
  2021, Virtual Event, February 2-9, 2021}}. \bibinfo{publisher}{{AAAI} Press},
  \bibinfo{pages}{3950--3957}.
\newblock
\href{https://doi.org/10.1609/AAAI.V35I5.16514}{doi:\nolinkurl{10.1609/AAAI.V35I5.16514}}


\bibitem[Borgwardt et~al\mbox{.}(2005)]%
        {ENZYMES}
\bibfield{author}{\bibinfo{person}{Karsten~M Borgwardt},
  \bibinfo{person}{Cheng~Soon Ong}, \bibinfo{person}{Stefan Sch{\"o}nauer},
  \bibinfo{person}{SVN Vishwanathan}, \bibinfo{person}{Alex~J Smola}, {and}
  \bibinfo{person}{Hans-Peter Kriegel}.} \bibinfo{year}{2005}\natexlab{}.
\newblock \showarticletitle{Protein function prediction via graph kernels}.
\newblock \bibinfo{journal}{\emph{Bioinformatics}} \bibinfo{volume}{21},
  \bibinfo{number}{suppl\_1} (\bibinfo{year}{2005}), \bibinfo{pages}{i47--i56}.
\newblock


\bibitem[Chien et~al\mbox{.}(2021)]%
        {GPR-GNN}
\bibfield{author}{\bibinfo{person}{Eli Chien}, \bibinfo{person}{Jianhao Peng},
  \bibinfo{person}{Pan Li}, {and} \bibinfo{person}{Olgica Milenkovic}.}
  \bibinfo{year}{2021}\natexlab{}.
\newblock \showarticletitle{Adaptive Universal Generalized PageRank Graph
  Neural Network}. In \bibinfo{booktitle}{\emph{9th International Conference on
  Learning Representations, {ICLR} 2021, Virtual Event, Austria, May 3-7,
  2021}}. \bibinfo{publisher}{OpenReview.net}.
\newblock
\urldef\tempurl%
\url{https://openreview.net/forum?id=n6jl7fLxrP}
\showURL{%
\tempurl}


\bibitem[Das et~al\mbox{.}(2018)]%
        {minerva}
\bibfield{author}{\bibinfo{person}{Rajarshi Das}, \bibinfo{person}{Shehzaad
  Dhuliawala}, \bibinfo{person}{Manzil Zaheer}, \bibinfo{person}{Luke Vilnis},
  \bibinfo{person}{Ishan Durugkar}, \bibinfo{person}{Akshay Krishnamurthy},
  \bibinfo{person}{Alex Smola}, {and} \bibinfo{person}{Andrew McCallum}.}
  \bibinfo{year}{2018}\natexlab{}.
\newblock \showarticletitle{Go for a Walk and Arrive at the Answer: Reasoning
  Over Paths in Knowledge Bases using Reinforcement Learning}. In
  \bibinfo{booktitle}{\emph{6th International Conference on Learning
  Representations, {ICLR} 2018, Vancouver, BC, Canada, April 30 - May 3, 2018,
  Conference Track Proceedings}}. \bibinfo{publisher}{OpenReview.net}.
\newblock
\urldef\tempurl%
\url{https://openreview.net/forum?id=Syg-YfWCW}
\showURL{%
\tempurl}


\bibitem[Dobson and Doig(2003)]%
        {Protein}
\bibfield{author}{\bibinfo{person}{Paul~D Dobson} {and}
  \bibinfo{person}{Andrew~J Doig}.} \bibinfo{year}{2003}\natexlab{}.
\newblock \showarticletitle{Distinguishing enzyme structures from non-enzymes
  without alignments}.
\newblock \bibinfo{journal}{\emph{Journal of molecular biology}}
  \bibinfo{volume}{330}, \bibinfo{number}{4} (\bibinfo{year}{2003}),
  \bibinfo{pages}{771--783}.
\newblock


\bibitem[Fang et~al\mbox{.}(2023)]%
        {GPF/GPF-plus}
\bibfield{author}{\bibinfo{person}{Taoran Fang}, \bibinfo{person}{Yunchao
  Zhang}, \bibinfo{person}{Yang Yang}, \bibinfo{person}{Chunping Wang}, {and}
  \bibinfo{person}{Lei Chen}.} \bibinfo{year}{2023}\natexlab{}.
\newblock \showarticletitle{Universal Prompt Tuning for Graph Neural Networks}.
  In \bibinfo{booktitle}{\emph{Advances in Neural Information Processing
  Systems 36: Annual Conference on Neural Information Processing Systems 2023,
  NeurIPS 2023, New Orleans, LA, USA, December 10 - 16, 2023}},
  \bibfield{editor}{\bibinfo{person}{Alice Oh}, \bibinfo{person}{Tristan
  Naumann}, \bibinfo{person}{Amir Globerson}, \bibinfo{person}{Kate Saenko},
  \bibinfo{person}{Moritz Hardt}, {and} \bibinfo{person}{Sergey Levine}}
  (Eds.).
\newblock


\bibitem[Hou et~al\mbox{.}(2022)]%
        {GraphMAE}
\bibfield{author}{\bibinfo{person}{Zhenyu Hou}, \bibinfo{person}{Xiao Liu},
  \bibinfo{person}{Yukuo Cen}, \bibinfo{person}{Yuxiao Dong},
  \bibinfo{person}{Hongxia Yang}, \bibinfo{person}{Chunjie Wang}, {and}
  \bibinfo{person}{Jie Tang}.} \bibinfo{year}{2022}\natexlab{}.
\newblock \showarticletitle{GraphMAE: Self-Supervised Masked Graph
  Autoencoders}. In \bibinfo{booktitle}{\emph{{KDD} '22: The 28th {ACM}
  {SIGKDD} Conference on Knowledge Discovery and Data Mining, Washington, DC,
  USA, August 14 - 18, 2022}}. \bibinfo{publisher}{{ACM}},
  \bibinfo{pages}{594--604}.
\newblock


\bibitem[Hu et~al\mbox{.}(2020)]%
        {dataOGB}
\bibfield{author}{\bibinfo{person}{Weihua Hu}, \bibinfo{person}{Matthias Fey},
  \bibinfo{person}{Marinka Zitnik}, \bibinfo{person}{Yuxiao Dong},
  \bibinfo{person}{Hongyu Ren}, \bibinfo{person}{Bowen Liu},
  \bibinfo{person}{Michele Catasta}, {and} \bibinfo{person}{Jure Leskovec}.}
  \bibinfo{year}{2020}\natexlab{}.
\newblock \showarticletitle{Open graph benchmark: Datasets for machine learning
  on graphs}.
\newblock \bibinfo{journal}{\emph{Advances in neural information processing
  systems}}  \bibinfo{volume}{33} (\bibinfo{year}{2020}),
  \bibinfo{pages}{22118--22133}.
\newblock


\bibitem[Kipf and Welling(2017)]%
        {GCN}
\bibfield{author}{\bibinfo{person}{Thomas~N. Kipf} {and} \bibinfo{person}{Max
  Welling}.} \bibinfo{year}{2017}\natexlab{}.
\newblock \showarticletitle{Semi-Supervised Classification with Graph
  Convolutional Networks}. In \bibinfo{booktitle}{\emph{5th International
  Conference on Learning Representations, {ICLR} 2017, Toulon, France, April
  24-26, 2017, Conference Track Proceedings}}.
  \bibinfo{publisher}{OpenReview.net}.
\newblock


\bibitem[Lai et~al\mbox{.}(2020)]%
        {policygnn}
\bibfield{author}{\bibinfo{person}{Kwei{-}Herng Lai}, \bibinfo{person}{Daochen
  Zha}, \bibinfo{person}{Kaixiong Zhou}, {and} \bibinfo{person}{Xia Hu}.}
  \bibinfo{year}{2020}\natexlab{}.
\newblock \showarticletitle{Policy-GNN: Aggregation Optimization for Graph
  Neural Networks}. In \bibinfo{booktitle}{\emph{{KDD} '20: The 26th {ACM}
  {SIGKDD} Conference on Knowledge Discovery and Data Mining, Virtual Event,
  CA, USA, August 23-27, 2020}}, \bibfield{editor}{\bibinfo{person}{Rajesh
  Gupta}, \bibinfo{person}{Yan Liu}, \bibinfo{person}{Jiliang Tang}, {and}
  \bibinfo{person}{B.~Aditya Prakash}} (Eds.). \bibinfo{publisher}{{ACM}},
  \bibinfo{pages}{461--471}.
\newblock
\href{https://doi.org/10.1145/3394486.3403088}{doi:\nolinkurl{10.1145/3394486.3403088}}


\bibitem[Liu et~al\mbox{.}(2026)]%
        {rgfm}
\bibfield{author}{\bibinfo{person}{Haokun Liu}, \bibinfo{person}{Zezhong Ding},
  {and} \bibinfo{person}{Xike Xie}.} \bibinfo{year}{2026}\natexlab{}.
\newblock \showarticletitle{Learning Graph Foundation Models on Riemannian
  Graph-of-Graphs}.
\newblock \bibinfo{journal}{\emph{CoRR}}  \bibinfo{volume}{abs/2605.09993}
  (\bibinfo{year}{2026}).
\newblock
\showeprint[arXiv]{2605.09993}
\href{https://doi.org/10.48550/ARXIV.2605.09993}{doi:\nolinkurl{10.48550/ARXIV.2605.09993}}


\bibitem[Liu et~al\mbox{.}(2023a)]%
        {liu2023towards}
\bibfield{author}{\bibinfo{person}{Jiawei Liu}, \bibinfo{person}{Cheng Yang},
  \bibinfo{person}{Zhiyuan Lu}, \bibinfo{person}{Junze Chen},
  \bibinfo{person}{Yibo Li}, \bibinfo{person}{Mengmei Zhang},
  \bibinfo{person}{Ting Bai}, \bibinfo{person}{Yuan Fang},
  \bibinfo{person}{Lichao Sun}, \bibinfo{person}{Philip~S Yu}, {et~al\mbox{.}}}
  \bibinfo{year}{2023}\natexlab{a}.
\newblock \showarticletitle{Towards graph foundation models: A survey and
  beyond}.
\newblock \bibinfo{journal}{\emph{arXiv preprint arXiv:2310.11829}}
  (\bibinfo{year}{2023}).
\newblock


\bibitem[Liu et~al\mbox{.}(2025)]%
        {KDEM}
\bibfield{author}{\bibinfo{person}{Lei Liu}, \bibinfo{person}{Xingyu Xia},
  \bibinfo{person}{Qianqian Xie}, \bibinfo{person}{Ben Liu},
  \bibinfo{person}{Wenjie Xu}, {and} \bibinfo{person}{Min Peng}.}
  \bibinfo{year}{2025}\natexlab{}.
\newblock \showarticletitle{Enhanced Expert Merging for Mixture-of-Experts in
  Graph Foundation Models}. In \bibinfo{booktitle}{\emph{The Thirty-ninth
  Annual Conference on Neural Information Processing Systems}}.
\newblock


\bibitem[Liu et~al\mbox{.}(2023b)]%
        {GraphPrompt}
\bibfield{author}{\bibinfo{person}{Zemin Liu}, \bibinfo{person}{Xingtong Yu},
  \bibinfo{person}{Yuan Fang}, {and} \bibinfo{person}{Xinming Zhang}.}
  \bibinfo{year}{2023}\natexlab{b}.
\newblock \showarticletitle{GraphPrompt: Unifying Pre-Training and Downstream
  Tasks for Graph Neural Networks}. In \bibinfo{booktitle}{\emph{Proceedings of
  the {ACM} Web Conference 2023, {WWW} 2023, Austin, TX, USA, 30 April 2023 - 4
  May 2023}}, \bibfield{editor}{\bibinfo{person}{Ying Ding},
  \bibinfo{person}{Jie Tang}, \bibinfo{person}{Juan~F. Sequeda},
  \bibinfo{person}{Lora Aroyo}, \bibinfo{person}{Carlos Castillo}, {and}
  \bibinfo{person}{Geert{-}Jan Houben}} (Eds.). \bibinfo{publisher}{{ACM}},
  \bibinfo{pages}{417--428}.
\newblock
\href{https://doi.org/10.1145/3543507.3583386}{doi:\nolinkurl{10.1145/3543507.3583386}}


\bibitem[Palomares et~al\mbox{.}(2021)]%
        {IF_App_recomend}
\bibfield{author}{\bibinfo{person}{Iv{\'{a}}n Palomares},
  \bibinfo{person}{Carlos Porcel}, \bibinfo{person}{Luiz Pizzato},
  \bibinfo{person}{Ido Guy}, {and} \bibinfo{person}{Enrique Herrera{-}Viedma}.}
  \bibinfo{year}{2021}\natexlab{}.
\newblock \showarticletitle{Reciprocal Recommender Systems: Analysis of
  state-of-art literature, challenges and opportunities towards social
  recommendation}.
\newblock \bibinfo{journal}{\emph{Inf. Fusion}}  \bibinfo{volume}{69}
  (\bibinfo{year}{2021}), \bibinfo{pages}{103--127}.
\newblock
\href{https://doi.org/10.1016/j.inffus.2020.12.001}{doi:\nolinkurl{10.1016/j.inffus.2020.12.001}}


\bibitem[Pei et~al\mbox{.}(2020)]%
        {Geom-GCN}
\bibfield{author}{\bibinfo{person}{Hongbin Pei}, \bibinfo{person}{Bingzhe Wei},
  \bibinfo{person}{Kevin~Chen{-}Chuan Chang}, \bibinfo{person}{Yu Lei}, {and}
  \bibinfo{person}{Bo Yang}.} \bibinfo{year}{2020}\natexlab{}.
\newblock \showarticletitle{Geom-GCN: Geometric Graph Convolutional Networks}.
  In \bibinfo{booktitle}{\emph{8th International Conference on Learning
  Representations, {ICLR} 2020, Addis Ababa, Ethiopia, April 26-30, 2020}}.
  \bibinfo{publisher}{OpenReview.net}.
\newblock


\bibitem[Schlichtkrull et~al\mbox{.}(2018)]%
        {r-gcn}
\bibfield{author}{\bibinfo{person}{Michael~Sejr Schlichtkrull},
  \bibinfo{person}{Thomas~N. Kipf}, \bibinfo{person}{Peter Bloem},
  \bibinfo{person}{Rianne van~den Berg}, \bibinfo{person}{Ivan Titov}, {and}
  \bibinfo{person}{Max Welling}.} \bibinfo{year}{2018}\natexlab{}.
\newblock \showarticletitle{Modeling Relational Data with Graph Convolutional
  Networks}. In \bibinfo{booktitle}{\emph{The Semantic Web - 15th International
  Conference, {ESWC} 2018, Heraklion, Crete, Greece, June 3-7, 2018,
  Proceedings}} \emph{(\bibinfo{series}{Lecture Notes in Computer Science},
  Vol.~\bibinfo{volume}{10843})}, \bibfield{editor}{\bibinfo{person}{Aldo
  Gangemi}, \bibinfo{person}{Roberto Navigli}, \bibinfo{person}{Maria{-}Esther
  Vidal}, \bibinfo{person}{Pascal Hitzler}, \bibinfo{person}{Rapha{\"{e}}l
  Troncy}, \bibinfo{person}{Laura Hollink}, \bibinfo{person}{Anna Tordai},
  {and} \bibinfo{person}{Mehwish Alam}} (Eds.). \bibinfo{publisher}{Springer},
  \bibinfo{pages}{593--607}.
\newblock
\href{https://doi.org/10.1007/978-3-319-93417-4\_38}{doi:\nolinkurl{10.1007/978-3-319-93417-4\_38}}


\bibitem[Shchur et~al\mbox{.}(2018)]%
        {amazon_computers_photo-coauthor_cs-physics}
\bibfield{author}{\bibinfo{person}{Oleksandr Shchur},
  \bibinfo{person}{Maximilian Mumme}, \bibinfo{person}{Aleksandar Bojchevski},
  {and} \bibinfo{person}{Stephan G{\"u}nnemann}.}
  \bibinfo{year}{2018}\natexlab{}.
\newblock \showarticletitle{Pitfalls of graph neural network evaluation}.
\newblock \bibinfo{journal}{\emph{arXiv preprint arXiv:1811.05868}}
  (\bibinfo{year}{2018}).
\newblock


\bibitem[Shervashidze et~al\mbox{.}(2011)]%
        {MUTAG_DD}
\bibfield{author}{\bibinfo{person}{Nino Shervashidze}, \bibinfo{person}{Pascal
  Schweitzer}, \bibinfo{person}{Erik~Jan van Leeuwen}, \bibinfo{person}{Kurt
  Mehlhorn}, {and} \bibinfo{person}{Karsten~M. Borgwardt}.}
  \bibinfo{year}{2011}\natexlab{}.
\newblock \showarticletitle{Weisfeiler-Lehman Graph Kernels}.
\newblock \bibinfo{journal}{\emph{J. Mach. Learn. Res.}}  \bibinfo{volume}{12}
  (\bibinfo{year}{2011}), \bibinfo{pages}{2539--2561}.
\newblock
\href{https://doi.org/10.5555/1953048.2078187}{doi:\nolinkurl{10.5555/1953048.2078187}}


\bibitem[Sun et~al\mbox{.}(2025)]%
        {RiemannGFM}
\bibfield{author}{\bibinfo{person}{Li Sun}, \bibinfo{person}{Zhenhao Huang},
  \bibinfo{person}{Suyang Zhou}, \bibinfo{person}{Qiqi Wan},
  \bibinfo{person}{Hao Peng}, {and} \bibinfo{person}{Philip~S. Yu}.}
  \bibinfo{year}{2025}\natexlab{}.
\newblock \showarticletitle{RiemannGFM: Learning a Graph Foundation Model from
  Riemannian Geometry}. In \bibinfo{booktitle}{\emph{Proceedings of the {ACM}
  on Web Conference 2025, {WWW} 2025, Sydney, NSW, Australia, 28 April 2025- 2
  May 2025}}, \bibfield{editor}{\bibinfo{person}{Guodong Long},
  \bibinfo{person}{Michale Blumestein}, \bibinfo{person}{Yi~Chang},
  \bibinfo{person}{Liane Lewin{-}Eytan}, \bibinfo{person}{Zi~Helen Huang},
  {and} \bibinfo{person}{Elad Yom{-}Tov}} (Eds.). \bibinfo{publisher}{{ACM}},
  \bibinfo{pages}{1154--1165}.
\newblock
\href{https://doi.org/10.1145/3696410.3714952}{doi:\nolinkurl{10.1145/3696410.3714952}}


\bibitem[Velickovic et~al\mbox{.}(2018)]%
        {GAT}
\bibfield{author}{\bibinfo{person}{Petar Velickovic}, \bibinfo{person}{Guillem
  Cucurull}, \bibinfo{person}{Arantxa Casanova}, \bibinfo{person}{Adriana
  Romero}, \bibinfo{person}{Pietro Li{\`{o}}}, {and} \bibinfo{person}{Yoshua
  Bengio}.} \bibinfo{year}{2018}\natexlab{}.
\newblock \showarticletitle{Graph Attention Networks}. In
  \bibinfo{booktitle}{\emph{6th International Conference on Learning
  Representations, {ICLR} 2018, Vancouver, BC, Canada, April 30 - May 3, 2018,
  Conference Track Proceedings}}. \bibinfo{publisher}{OpenReview.net}.
\newblock


\bibitem[Velickovic et~al\mbox{.}(2019)]%
        {DGI}
\bibfield{author}{\bibinfo{person}{Petar Velickovic}, \bibinfo{person}{William
  Fedus}, \bibinfo{person}{William~L. Hamilton}, \bibinfo{person}{Pietro
  Li{\`{o}}}, \bibinfo{person}{Yoshua Bengio}, {and} \bibinfo{person}{R.~Devon
  Hjelm}.} \bibinfo{year}{2019}\natexlab{}.
\newblock \showarticletitle{Deep {G}raph {I}nfomax}. In
  \bibinfo{booktitle}{\emph{7th International Conference on Learning
  Representations, {ICLR} 2019, New Orleans, LA, USA, May 6-9, 2019}}.
  \bibinfo{publisher}{OpenReview.net}.
\newblock


\bibitem[Vlaic et~al\mbox{.}(2018)]%
        {appBio}
\bibfield{author}{\bibinfo{person}{Sebastian Vlaic}, \bibinfo{person}{Theresia
  Conrad}, \bibinfo{person}{Christian Tokarski-Schnelle}, \bibinfo{person}{Mika
  Gustafsson}, \bibinfo{person}{Uta Dahmen}, \bibinfo{person}{Reinhard Guthke},
  {and} \bibinfo{person}{Stefan Schuster}.} \bibinfo{year}{2018}\natexlab{}.
\newblock \showarticletitle{ModuleDiscoverer: Identification of regulatory
  modules in protein-protein interaction networks}.
\newblock \bibinfo{journal}{\emph{Scientific reports}} \bibinfo{volume}{8},
  \bibinfo{number}{1} (\bibinfo{year}{2018}), \bibinfo{pages}{433}.
\newblock


\bibitem[Wang et~al\mbox{.}(2024)]%
        {GFT}
\bibfield{author}{\bibinfo{person}{Zehong Wang}, \bibinfo{person}{Zheyuan
  Zhang}, \bibinfo{person}{Nitesh~V. Chawla}, \bibinfo{person}{Chuxu Zhang},
  {and} \bibinfo{person}{Yanfang Ye}.} \bibinfo{year}{2024}\natexlab{}.
\newblock \showarticletitle{{GFT:} Graph Foundation Model with Transferable
  Tree Vocabulary}. In \bibinfo{booktitle}{\emph{Advances in Neural Information
  Processing Systems 38: Annual Conference on Neural Information Processing
  Systems 2024, NeurIPS 2024, Vancouver, BC, Canada, December 10 - 15, 2024}},
  \bibfield{editor}{\bibinfo{person}{Amir Globersons}, \bibinfo{person}{Lester
  Mackey}, \bibinfo{person}{Danielle Belgrave}, \bibinfo{person}{Angela Fan},
  \bibinfo{person}{Ulrich Paquet}, \bibinfo{person}{Jakub~M. Tomczak}, {and}
  \bibinfo{person}{Cheng Zhang}} (Eds.).
\newblock


\bibitem[Wu et~al\mbox{.}(2019)]%
        {appFinance}
\bibfield{author}{\bibinfo{person}{Shu Wu}, \bibinfo{person}{Yuyuan Tang},
  \bibinfo{person}{Yanqiao Zhu}, \bibinfo{person}{Liang Wang},
  \bibinfo{person}{Xing Xie}, {and} \bibinfo{person}{Tieniu Tan}.}
  \bibinfo{year}{2019}\natexlab{}.
\newblock \showarticletitle{Session-Based Recommendation with Graph Neural
  Networks}. In \bibinfo{booktitle}{\emph{The Thirty-Third {AAAI} Conference on
  Artificial Intelligence, {AAAI} 2019, The Thirty-First Innovative
  Applications of Artificial Intelligence Conference, {IAAI} 2019, The Ninth
  {AAAI} Symposium on Educational Advances in Artificial Intelligence, {EAAI}
  2019, Honolulu, Hawaii, USA, January 27 - February 1, 2019}}.
  \bibinfo{publisher}{{AAAI} Press}, \bibinfo{pages}{346--353}.
\newblock


\bibitem[Xia et~al\mbox{.}(2024)]%
        {opengraph}
\bibfield{author}{\bibinfo{person}{Lianghao Xia}, \bibinfo{person}{Ben Kao},
  {and} \bibinfo{person}{Chao Huang}.} \bibinfo{year}{2024}\natexlab{}.
\newblock \showarticletitle{OpenGraph: Towards Open Graph Foundation Models}.
  In \bibinfo{booktitle}{\emph{Findings of the Association for Computational
  Linguistics: {EMNLP} 2024, Miami, Florida, USA, November 12-16, 2024}}
  \emph{(\bibinfo{series}{Findings of {ACL}}, Vol.~\bibinfo{volume}{{EMNLP}
  2024})}, \bibfield{editor}{\bibinfo{person}{Yaser Al{-}Onaizan},
  \bibinfo{person}{Mohit Bansal}, {and} \bibinfo{person}{Yun{-}Nung Chen}}
  (Eds.). \bibinfo{publisher}{Association for Computational Linguistics},
  \bibinfo{pages}{2365--2379}.
\newblock
\href{https://doi.org/10.18653/V1/2024.FINDINGS-EMNLP.132}{doi:\nolinkurl{10.18653/V1/2024.FINDINGS-EMNLP.132}}


\bibitem[Xiong et~al\mbox{.}(2017)]%
        {deeppath}
\bibfield{author}{\bibinfo{person}{Wenhan Xiong}, \bibinfo{person}{Thien
  Hoang}, {and} \bibinfo{person}{William~Yang Wang}.}
  \bibinfo{year}{2017}\natexlab{}.
\newblock \showarticletitle{DeepPath: {A} Reinforcement Learning Method for
  Knowledge Graph Reasoning}. In \bibinfo{booktitle}{\emph{Proceedings of the
  2017 Conference on Empirical Methods in Natural Language Processing, {EMNLP}
  2017, Copenhagen, Denmark, September 9-11, 2017}},
  \bibfield{editor}{\bibinfo{person}{Martha Palmer}, \bibinfo{person}{Rebecca
  Hwa}, {and} \bibinfo{person}{Sebastian Riedel}} (Eds.).
  \bibinfo{publisher}{Association for Computational Linguistics},
  \bibinfo{pages}{564--573}.
\newblock
\href{https://doi.org/10.18653/V1/D17-1060}{doi:\nolinkurl{10.18653/V1/D17-1060}}


\bibitem[Xue et~al\mbox{.}(2021)]%
        {gdpnet}
\bibfield{author}{\bibinfo{person}{Fuzhao Xue}, \bibinfo{person}{Aixin Sun},
  \bibinfo{person}{Hao Zhang}, {and} \bibinfo{person}{Eng~Siong Chng}.}
  \bibinfo{year}{2021}\natexlab{}.
\newblock \showarticletitle{GDPNet: Refining Latent Multi-View Graph for
  Relation Extraction}. In \bibinfo{booktitle}{\emph{Thirty-Fifth {AAAI}
  Conference on Artificial Intelligence, {AAAI} 2021, Thirty-Third Conference
  on Innovative Applications of Artificial Intelligence, {IAAI} 2021, The
  Eleventh Symposium on Educational Advances in Artificial Intelligence, {EAAI}
  2021, Virtual Event, February 2-9, 2021}}. \bibinfo{publisher}{{AAAI} Press},
  \bibinfo{pages}{14194--14202}.
\newblock
\href{https://doi.org/10.1609/AAAI.V35I16.17670}{doi:\nolinkurl{10.1609/AAAI.V35I16.17670}}


\bibitem[Yanardag and Vishwanathan(2015)]%
        {COLLAB_IMDB}
\bibfield{author}{\bibinfo{person}{Pinar Yanardag} {and}
  \bibinfo{person}{S.~V.~N. Vishwanathan}.} \bibinfo{year}{2015}\natexlab{}.
\newblock \showarticletitle{Deep Graph Kernels}. In
  \bibinfo{booktitle}{\emph{Proceedings of the 21th {ACM} {SIGKDD}
  International Conference on Knowledge Discovery and Data Mining, Sydney, NSW,
  Australia, August 10-13, 2015}}, \bibfield{editor}{\bibinfo{person}{Longbing
  Cao}, \bibinfo{person}{Chengqi Zhang}, \bibinfo{person}{Thorsten Joachims},
  \bibinfo{person}{Geoffrey~I. Webb}, \bibinfo{person}{Dragos~D. Margineantu},
  {and} \bibinfo{person}{Graham Williams}} (Eds.). \bibinfo{publisher}{{ACM}},
  \bibinfo{pages}{1365--1374}.
\newblock
\href{https://doi.org/10.1145/2783258.2783417}{doi:\nolinkurl{10.1145/2783258.2783417}}


\bibitem[Yang et~al\mbox{.}(2016)]%
        {citation_cora_citeseer_pubmed}
\bibfield{author}{\bibinfo{person}{Zhilin Yang}, \bibinfo{person}{William
  Cohen}, {and} \bibinfo{person}{Ruslan Salakhudinov}.}
  \bibinfo{year}{2016}\natexlab{}.
\newblock \showarticletitle{Revisiting semi-supervised learning with graph
  embeddings}. In \bibinfo{booktitle}{\emph{International conference on machine
  learning}}. PMLR, \bibinfo{pages}{40--48}.
\newblock


\bibitem[Yang et~al\mbox{.}(2025)]%
        {GraphLoRA}
\bibfield{author}{\bibinfo{person}{Zhe{-}Rui Yang}, \bibinfo{person}{Jindong
  Han}, \bibinfo{person}{Chang{-}Dong Wang}, {and} \bibinfo{person}{Hao Liu}.}
  \bibinfo{year}{2025}\natexlab{}.
\newblock \showarticletitle{GraphLoRA: Structure-Aware Contrastive Low-Rank
  Adaptation for Cross-Graph Transfer Learning}. In
  \bibinfo{booktitle}{\emph{Proceedings of the 31st {ACM} {SIGKDD} Conference
  on Knowledge Discovery and Data Mining, V.1, {KDD} 2025, Toronto, ON, Canada,
  August 3-7, 2025}}, \bibfield{editor}{\bibinfo{person}{Yizhou Sun},
  \bibinfo{person}{Flavio Chierichetti}, \bibinfo{person}{Hady~W. Lauw},
  \bibinfo{person}{Claudia Perlich}, \bibinfo{person}{Wee~Hyong Tok}, {and}
  \bibinfo{person}{Andrew Tomkins}} (Eds.). \bibinfo{publisher}{{ACM}},
  \bibinfo{pages}{1785--1796}.
\newblock
\href{https://doi.org/10.1145/3690624.3709186}{doi:\nolinkurl{10.1145/3690624.3709186}}


\bibitem[You et~al\mbox{.}(2020)]%
        {GRAPHCL}
\bibfield{author}{\bibinfo{person}{Yuning You}, \bibinfo{person}{Tianlong
  Chen}, \bibinfo{person}{Yongduo Sui}, \bibinfo{person}{Ting Chen},
  \bibinfo{person}{Zhangyang Wang}, {and} \bibinfo{person}{Yang Shen}.}
  \bibinfo{year}{2020}\natexlab{}.
\newblock \showarticletitle{Graph Contrastive Learning with Augmentations}. In
  \bibinfo{booktitle}{\emph{Advances in Neural Information Processing Systems
  33: Annual Conference on Neural Information Processing Systems 2020, NeurIPS
  2020, December 6-12, 2020, virtual}}, \bibfield{editor}{\bibinfo{person}{Hugo
  Larochelle}, \bibinfo{person}{Marc'Aurelio Ranzato}, \bibinfo{person}{Raia
  Hadsell}, \bibinfo{person}{Maria{-}Florina Balcan}, {and}
  \bibinfo{person}{Hsuan{-}Tien Lin}} (Eds.).
\newblock


\bibitem[Yu et~al\mbox{.}(2025)]%
        {SAMGPT}
\bibfield{author}{\bibinfo{person}{Xingtong Yu}, \bibinfo{person}{Zechuan
  Gong}, \bibinfo{person}{Chang Zhou}, \bibinfo{person}{Yuan Fang}, {and}
  \bibinfo{person}{Hui Zhang}.} \bibinfo{year}{2025}\natexlab{}.
\newblock \showarticletitle{{SAMGPT:} Text-free Graph Foundation Model for
  Multi-domain Pre-training and Cross-domain Adaptation}. In
  \bibinfo{booktitle}{\emph{Proceedings of the {ACM} on Web Conference 2025,
  {WWW} 2025, Sydney, NSW, Australia, 28 April 2025- 2 May 2025}},
  \bibfield{editor}{\bibinfo{person}{Guodong Long}, \bibinfo{person}{Michale
  Blumestein}, \bibinfo{person}{Yi~Chang}, \bibinfo{person}{Liane
  Lewin{-}Eytan}, \bibinfo{person}{Zi~Helen Huang}, {and} \bibinfo{person}{Elad
  Yom{-}Tov}} (Eds.). \bibinfo{publisher}{{ACM}}, \bibinfo{pages}{1142--1153}.
\newblock
\href{https://doi.org/10.1145/3696410.3714828}{doi:\nolinkurl{10.1145/3696410.3714828}}


\bibitem[Yu et~al\mbox{.}(2024)]%
        {mdgpt}
\bibfield{author}{\bibinfo{person}{Xingtong Yu}, \bibinfo{person}{Chang Zhou},
  \bibinfo{person}{Yuan Fang}, {and} \bibinfo{person}{Xinming Zhang}.}
  \bibinfo{year}{2024}\natexlab{}.
\newblock \showarticletitle{Text-free multi-domain graph pre-training: Toward
  graph foundation models}.
\newblock \bibinfo{journal}{\emph{arXiv preprint arXiv:2405.13934}}
  (\bibinfo{year}{2024}).
\newblock


\bibitem[Yuan et~al\mbox{.}(2025)]%
        {BRIDGE}
\bibfield{author}{\bibinfo{person}{Haonan Yuan}, \bibinfo{person}{Qingyun Sun},
  \bibinfo{person}{Junhua Shi}, \bibinfo{person}{Xingcheng Fu},
  \bibinfo{person}{Bryan Hooi}, \bibinfo{person}{Jianxin Li}, {and}
  \bibinfo{person}{Philip~S. Yu}.} \bibinfo{year}{2025}\natexlab{}.
\newblock \showarticletitle{How Much Can Transfer? {BRIDGE}: Bounded
  Multi-Domain Graph Foundation Model with Generalization Guarantees}. In
  \bibinfo{booktitle}{\emph{Forty-second International Conference on Machine
  Learning}}.
\newblock


\bibitem[Zhao et~al\mbox{.}(2024)]%
        {GCOPE}
\bibfield{author}{\bibinfo{person}{Haihong Zhao}, \bibinfo{person}{Aochuan
  Chen}, \bibinfo{person}{Xiangguo Sun}, \bibinfo{person}{Hong Cheng}, {and}
  \bibinfo{person}{Jia Li}.} \bibinfo{year}{2024}\natexlab{}.
\newblock \showarticletitle{All in One and One for All: {A} Simple yet
  Effective Method towards Cross-domain Graph Pretraining}. In
  \bibinfo{booktitle}{\emph{Proceedings of the 30th {ACM} {SIGKDD} Conference
  on Knowledge Discovery and Data Mining, {KDD} 2024, Barcelona, Spain, August
  25-29, 2024}}, \bibfield{editor}{\bibinfo{person}{Ricardo Baeza{-}Yates}
  {and} \bibinfo{person}{Francesco Bonchi}} (Eds.). \bibinfo{publisher}{{ACM}},
  \bibinfo{pages}{4443--4454}.
\newblock
\href{https://doi.org/10.1145/3637528.3671913}{doi:\nolinkurl{10.1145/3637528.3671913}}


\bibitem[Zhao et~al\mbox{.}(2026)]%
        {TIG}
\bibfield{author}{\bibinfo{person}{Jitao Zhao}, \bibinfo{person}{Yi Wang},
  \bibinfo{person}{Yawen Li}, \bibinfo{person}{Dongxiao He},
  \bibinfo{person}{Di Jin}, \bibinfo{person}{Zhiyong Feng}, {and}
  \bibinfo{person}{Weixiong Zhang}.} \bibinfo{year}{2026}\natexlab{}.
\newblock \showarticletitle{Towards Graph Foundation Model: Node Feature
  Transfer Invariant Modeling on General Graphs}. In
  \bibinfo{booktitle}{\emph{Proceedings of the ACM Web Conference 2026}}
  (United Arab Emirates) \emph{(\bibinfo{series}{WWW '26})}.
  \bibinfo{publisher}{Association for Computing Machinery},
  \bibinfo{address}{New York, NY, USA}, \bibinfo{pages}{810–821}.
\newblock
\showISBNx{9798400723070}
\href{https://doi.org/10.1145/3774904.3792236}{doi:\nolinkurl{10.1145/3774904.3792236}}


\bibitem[Zhao et~al\mbox{.}(2025b)]%
        {GraphAny}
\bibfield{author}{\bibinfo{person}{Jianan Zhao}, \bibinfo{person}{Zhaocheng
  Zhu}, \bibinfo{person}{Mikhail Galkin}, \bibinfo{person}{Hesham Mostafa},
  \bibinfo{person}{Michael~M. Bronstein}, {and} \bibinfo{person}{Jian Tang}.}
  \bibinfo{year}{2025}\natexlab{b}.
\newblock \showarticletitle{Fully-inductive Node Classification on Arbitrary
  Graphs}. In \bibinfo{booktitle}{\emph{The Thirteenth International Conference
  on Learning Representations, {ICLR} 2025, Singapore, April 24-28, 2025}}.
  \bibinfo{publisher}{OpenReview.net}.
\newblock


\bibitem[Zhao et~al\mbox{.}(2025a)]%
        {graphgpt}
\bibfield{author}{\bibinfo{person}{Qifang Zhao}, \bibinfo{person}{Weidong Ren},
  \bibinfo{person}{Tianyu Li}, \bibinfo{person}{Hong Liu},
  \bibinfo{person}{Xingsheng He}, {and} \bibinfo{person}{Xiaoxiao Xu}.}
  \bibinfo{year}{2025}\natexlab{a}.
\newblock \showarticletitle{GraphGPT: Generative Pre-trained Graph Eulerian
  Transformer}. In \bibinfo{booktitle}{\emph{Forty-second International
  Conference on Machine Learning, {ICML} 2025, Vancouver, BC, Canada, July
  13-19, 2025}} \emph{(\bibinfo{series}{Proceedings of Machine Learning
  Research}, Vol.~\bibinfo{volume}{267})},
  \bibfield{editor}{\bibinfo{person}{Aarti Singh}, \bibinfo{person}{Maryam
  Fazel}, \bibinfo{person}{Daniel Hsu}, \bibinfo{person}{Simon
  Lacoste{-}Julien}, \bibinfo{person}{Felix Berkenkamp}, \bibinfo{person}{Tegan
  Maharaj}, \bibinfo{person}{Kiri Wagstaff}, {and} \bibinfo{person}{Jerry Zhu}}
  (Eds.). \bibinfo{publisher}{{PMLR} / OpenReview.net}.
\newblock
\urldef\tempurl%
\url{https://proceedings.mlr.press/v267/zhao25r.html}
\showURL{%
\tempurl}


\bibitem[Zhu et~al\mbox{.}(2024)]%
        {GraphControl}
\bibfield{author}{\bibinfo{person}{Yun Zhu}, \bibinfo{person}{Yaoke Wang},
  \bibinfo{person}{Haizhou Shi}, \bibinfo{person}{Zhenshuo Zhang},
  \bibinfo{person}{Dian Jiao}, {and} \bibinfo{person}{Siliang Tang}.}
  \bibinfo{year}{2024}\natexlab{}.
\newblock \showarticletitle{GraphControl: Adding Conditional Control to
  Universal Graph Pre-trained Models for Graph Domain Transfer Learning}. In
  \bibinfo{booktitle}{\emph{Proceedings of the ACM Web Conference 2024}}
  \emph{(\bibinfo{series}{WWW ’24})}. \bibinfo{publisher}{ACM},
  \bibinfo{pages}{539–550}.
\newblock
\href{https://doi.org/10.1145/3589334.3645439}{doi:\nolinkurl{10.1145/3589334.3645439}}


\end{thebibliography}

\appendix

\section{Mechanism-Level Positioning}
\label{app:mechanism_positioning}

To further clarify the distinction between AgentGFM and existing graph
propagation paradigms, we compare their primary control objects,
optimization signals and interaction mechanisms in
Table~\ref{tab:propagation_positioning}. The comparison focuses on how
information-flow execution is determined rather than on specific model
architectures.

\begin{table*}[t]
\centering
\caption{Mechanism-level positioning of AgentGFM among graph propagation
paradigms. Opt. Signal denotes the objective used for parameter learning,
while Interaction Feedback denotes contextual signals used to condition
subsequent information-flow decisions.}
\Description{
A comparison of six graph propagation paradigms in terms of source
control, signal control, depth control, optimization signal and
interaction feedback. AgentGFM differs from the other paradigms by
combining node- and edge-level source control, low- and high-frequency
signal selection, node-wise halting, pretraining objectives and explicit
prediction-observation feedback.
}
\label{tab:propagation_positioning}

\begin{tabular}{lccccc}
\toprule
Paradigm
& Source Control
& Signal Control
& Depth Control
& Opt. Signal
& Interaction Feedback \\
\midrule

Fixed MP
& Architecture-defined
& Single
& Fixed
& Task loss
& None \\

Attention MP
& Attention-weighted
& Single
& Fixed
& Task loss
& None \\

Multi-hop MP
& Hop-defined
& Single
& Preset
& Task loss
& None \\

Adaptive GNNs
& Learned
& Limited
& Adaptive
& Task loss
& No explicit feedback \\

Component-Adaptive GFMs
& Model/expert
& Model/expert
& Fixed/expert
& Pretraining
& None \\

\textbf{AgentGFM}
& \textbf{Node/edge}
& \textbf{Low/high}
& \textbf{Node-wise halt}
& \textbf{Pretraining}
& \textbf{Pred.--obs.} \\

\bottomrule
\end{tabular}
\end{table*}
Table~\ref{tab:propagation_positioning} distinguishes existing
paradigms according to the primary object controlled during
information-flow execution. Fixed and multi-hop message passing specify
information sources and propagation ranges mainly through the model
architecture. Attention-based methods adapt neighbor importance, while
adaptive GNNs may further adjust neighborhood selection or propagation
depth. However, these methods generally optimize task-specific
propagation decisions without explicitly using contextual observations
to correct node states and condition subsequent actions.

Component-adaptive GFMs mainly adapt prompts, experts, structural
encodings, or model components for cross-graph transfer, while the
underlying information-flow execution remains largely predetermined.
In contrast, AgentGFM treats the node-specific information-flow
trajectory as the adaptive object. Each node controls source reception,
signal-channel selection and node-wise halting and uses
prediction--observation feedback to correct its state and condition
subsequent interactions.

\section{Experimental Settings}
\subsection{Datasets}
\label{app:datasets}
We evaluate node-level performance on a diverse set of benchmark graphs.
Citation networks include Cora, CiteSeer and PubMed~\cite{citation_cora_citeseer_pubmed}.
Web page networks consist of Texas, Cornell and Wisconsin~\cite{Geom-GCN}.
Wikipedia topic graphs include Chameleon and Squirrel~\cite{Geom-GCN}.
E-commerce graphs include  Photo  and Computers~\cite{amazon_computers_photo-coauthor_cs-physics}.
We additionally consider the large-scale ogbn-products and ogbn-arxiv datasets from the Open Graph Benchmark~\cite{dataOGB} and Physics~\cite{amazon_computers_photo-coauthor_cs-physics}
\begin{table}[t]
\centering
\caption{Overview of node classification datasets used in the experiments.}
\Description{
Statistics of thirteen node classification datasets, including the
numbers of nodes, edges, input features and classes.
}
\label{tab:node_dataset_description}

\resizebox{\columnwidth}{!}{
\begin{tabular}{lcccc}
\toprule
\textbf{Dataset}
& \textbf{\#Nodes}
& \textbf{\#Edges}
& \textbf{\#Features}
& \textbf{\#Classes} \\
\midrule
Texas
& 183
& 325
& 1,703
& 5 \\

Cornell
& 183
& 298
& 1,703
& 5 \\

Wisconsin
& 251
& 512
& 1,703
& 5 \\

Cora
& 2,708
& 10,556
& 1,433
& 7 \\

CiteSeer
& 3,327
& 9,104
& 3,703
& 6 \\

PubMed
& 19,717
& 88,648
& 500
& 3 \\

Chameleon
& 2,277
& 36,101
& 2,325
& 5 \\

Squirrel
& 5,201
& 217,073
& 2,089
& 5 \\

Computers
& 13,752
& 491,722
& 767
& 10 \\

Photo
& 7,650
& 238,162
& 745
& 8 \\

Physics
& 34,493
& 495,924
& 8,415
& 5 \\

ogbn-arxiv
& 169,343
& 1,166,243
& 128
& 40 \\

ogbn-products
& 2,449,029
& 123,718,280
& 100
& 47 \\
\bottomrule
\end{tabular}
}
\end{table}

For graph-level tasks, we use molecular and biological datasets including MUTAG, DD~\cite{MUTAG_DD}, ENZYMES~\cite{ENZYMES} and PROTEINS~\cite{Protein}.
We further include social network datasets IMDB-BINARY~\cite{COLLAB_IMDB}.
All datasets are commonly used benchmarks in graph representation learning and cover a wide range of graph sizes, structures and homophily characteristics.
For graph classification datasets from TUDataset, if the original graphs do not contain node features, we construct node features as follows. If node labels are available, we use one-hot encodings of node labels as node features. Otherwise, we use one-hot encodings of node degrees, where the one-hot dimension is determined by the maximum label value or maximum degree across the entire dataset.
If the original dataset provides node features $x$, we directly use them.

\begin{table}[t]
\centering
\caption{Overview of graph classification datasets used in the experiments.
Avg. Nodes and Avg. Edges denote the average numbers of nodes and edges
per graph, respectively.}
\Description{
Statistics of five graph classification datasets, including the number
of graphs, average graph size, input feature dimension and number of
classes. IMDB-BINARY contains no original node attributes.
}
\label{tab:graph_dataset_description}

\resizebox{\columnwidth}{!}{
\begin{tabular}{lccccc}
\toprule
\textbf{Dataset}
& \textbf{\#Graphs}
& \textbf{Avg. Nodes}
& \textbf{Avg. Edges}
& \textbf{\#Features}
& \textbf{\#Classes} \\
\midrule
MUTAG
& 188
& 17.9
& 39.6
& 7
& 2 \\

ENZYMES
& 600
& 32.6
& 124.3
& 3
& 6 \\

PROTEINS
& 1,113
& 39.1
& 145.6
& 3
& 2 \\

IMDB-BINARY
& 1,000
& 19.8
& 193.1
& --
& 2 \\

DD
& 1,178
& 284.3
& 1,431.3
& 89
& 2 \\
\bottomrule
\end{tabular}
}
\end{table}

\begin{table*}[t]
\caption{Full component ablation results on cross-domain 1-shot node
classification. Source-Reception Control corresponds to $\gamma$,
Channel Selection corresponds to $\rho$, Feedback denotes
prediction--observation feedback and Gain-Aware Halting denotes the
node-wise halting mechanism. The best result on each dataset is
highlighted in bold. Avg. Rank denotes the average rank across all
datasets, where a lower value is better.}
\label{tab:full_ablation}
\centering
\resizebox{\textwidth}{!}{
\begin{tabular}{lccccccccccc}
\toprule
Method
& Cora
& CiteSeer
& PubMed
& Computers
& Photo
& Texas
& Wisconsin
& Cornell
& Chameleon
& Squirrel
& Avg. Rank \\
\midrule
Full
& \textbf{0.5194}
& \textbf{0.4432}
& \textbf{0.5296}
& \textbf{0.5593}
& \textbf{0.6748}
& \textbf{0.4221}
& 0.4457
& \textbf{0.3983}
& 0.3054
& \textbf{0.2420}
& \textbf{1.2} \\

w/o Source-Reception Control
& 0.5143
& 0.4369
& 0.5238
& 0.5309
& 0.6631
& 0.3949
& 0.4300
& 0.3646
& 0.2930
& 0.2365
& 2.9 \\

w/o Channel Selection
& 0.4962
& 0.4333
& 0.4932
& 0.5152
& 0.6416
& 0.3382
& 0.4063
& 0.3474
& 0.2948
& 0.2342
& 3.9 \\

w/o Feedback
& 0.3408
& 0.3614
& 0.4220
& 0.4853
& 0.3637
& 0.3211
& 0.4251
& 0.3899
& 0.2990
& 0.2376
& 4.2 \\

w/o Gain-Aware Halting
& 0.5044
& 0.4195
& 0.4772
& 0.5061
& 0.5857
& 0.4189
& \textbf{0.4496}
& 0.3720
& \textbf{0.3073}
& 0.2410
& 2.8 \\
\bottomrule
\end{tabular}
}
\end{table*}

\subsection{Implementation Details}

\begin{table}[t]
\caption{Default hyperparameter settings of AgentGFM.}
\label{tab:implementation_hyperparameters}
\centering
\begin{tabular}{lc}
\toprule
\textbf{Hyperparameter} & \textbf{Value} \\
\midrule
Shared attribute dimension & 50 \\
Hidden dimension & 128 \\
Interaction rounds $R$ & 2 \\
Maximum rollout horizon $T_{\max}$ & 5 \\
Dropout rate & 0.10 \\
Masking ratio & 0.35 \\
Learning rate & $5\times10^{-4}$ \\
Weight decay & $5\times10^{-4}$ \\
Training epochs & 500 \\
\midrule
Halt threshold $\theta_0$ & 0.55 \\
High-frequency channel coefficient $\lambda_{\mathrm{high}}$ & 0.30 \\
\midrule
$\lambda_{\mathrm{mae}}$ & 1.00 \\
$\lambda_{\mathrm{pred}}$ & 0.50 \\
$\lambda_{\gamma}$ & 0.001 \\
$\lambda_{\mu}$ & 0.001 \\
$\lambda_a$ & 0.001 \\
\midrule
Number of 1-shot runs & 100 \\
\bottomrule
\end{tabular}
\end{table}

We implement AgentGFM in PyTorch and use the same default configuration
across all datasets unless otherwise specified. Node attributes are
projected into a shared 50-dimensional space and the hidden dimension
is set to 128. The model contains two
predict--act--observe--correct interaction rounds, with the maximum
rollout horizon set to $T_{\max}=5$. We train AgentGFM for 500 epochs
using a learning rate of $5\times10^{-4}$, a weight decay of
$5\times10^{-4}$, a dropout rate of 0.1 and a masking ratio of 0.35.
The complete implementation and hyperparameter settings are summarized
in Table~\ref{tab:implementation_hyperparameters}. For 1-shot
evaluation, we repeat the sampling process 100 times and report the
mean accuracy and standard deviation.

\subsection{Information-Flow Policy and Regularization Details}
\label{app:information_flow_policy}

This section provides additional details of the node-level
information-flow policy and the associated regularizers used in
AgentGFM. The policy is shared across nodes and graphs, while its
source-reception, channel-selection, forwarding-budget and halting
decisions are computed for each local rollout trajectory.

\subsubsection{Information-Flow Policy}

\paragraph{Source-reception policy.}
For each original edge $(u,i)\in\mathcal E$, the source-reception
strength is computed from carrier compatibility,
prediction--observation feedback and the source forwarding budget:
\begin{equation}
s_{u\rightarrow i}^{\gamma,t}
=
\frac{
\left(
\mathbf W_q^\gamma \mathbf c_u^t
\right)^\top
\left(
\mathbf W_k^\gamma \mathbf c_i^t
\right)
}{
\sqrt r
}
+
g_\gamma
\left(
\left[
\epsilon_u
\Vert
\epsilon_i
\Vert
b_u^t
\right]
\right),
\end{equation}
\begin{equation}
\gamma_{u\rightarrow i}^{t}
=
\sigma
\left(
s_{u\rightarrow i}^{\gamma,t}
\right),
\end{equation}
where $\epsilon_u$ and $\epsilon_i$ are prediction--observation
feedback cues and $b_u^t$ is the source forwarding budget.
The resulting $\gamma_{u\rightarrow i}^{t}$ measures the
source-reception strength from node $v_u$ to node $v_i$ under the
current rollout state.

\paragraph{Channel-selection policy.}
The channel-selection gate determines how low- and high-frequency
information are combined along each edge. We compute
\begin{equation}
\rho_{u\rightarrow i}^{t}
=
\sigma
\left(
\frac{
s_{u\rightarrow i}^{\rho,t}
+
\beta_{\mathrm{smooth}}
}{
\tau_\rho
}
\right),
\end{equation}
where
\begin{align}
s_{u\rightarrow i}^{\rho,t}
=
&
\frac{
\left(
\mathbf W_q^\rho \mathbf c_u^t
\right)^\top
\left(
\mathbf W_k^\rho \mathbf c_i^t
\right)
}{
\sqrt r
}
\nonumber\\
&+
g_\rho
\left(
\left[
s_{ui}^t
\Vert
1-s_{ui}^t
\Vert
\epsilon_u
\Vert
\epsilon_i
\Vert
b_u^t
\right]
\right).
\end{align}
Here, $s_{ui}^t$ denotes the cosine-similarity diagnostic between
the source and target carrier states. The smoothness bias
$\beta_{\mathrm{smooth}}$ and temperature $\tau_\rho$ control the
preference and sharpness of the low-/high-frequency interpolation.
The final message is
\begin{equation}
\mathbf m_{u\rightarrow i}^{t}
=
\rho_{u\rightarrow i}^{t}
\mathbf m_{u,\mathrm{low}}^{t}
+
\left(
1-\rho_{u\rightarrow i}^{t}
\right)
\lambda_{\mathrm{high}}
\mathbf m_{u,\mathrm{high}}^{t}.
\end{equation}
A larger $\rho_{u\rightarrow i}^{t}$ assigns greater weight to
low-frequency contextual information, whereas a smaller value assigns
greater weight to high-frequency residual information.

\paragraph{Forwarding-budget update.}
The forwarding budget represents the current capacity of a node to
forward contextual information rather than a monotonically consumed
resource. It is updated according to the effective information mass
received at the current rollout step:
\begin{equation}
\widetilde b_i^{t+1}
=
\mathcal T_{b_{\min}}
\left[
\operatorname{clip}
\left(
\eta_b
\sum_{u\in\mathcal N(i)}
\mu_{u\rightarrow i}^{t},
0,1
\right)
\right],
\end{equation}
where $\eta_b$ is the budget coefficient and
$\mathcal T_{b_{\min}}(\cdot)$ sets values below $b_{\min}$ to zero.
The budget state is updated according to the current activity status:
\begin{equation}
b_i^{t+1}
=
a_i^t
\widetilde b_i^{t+1}
+
\left(
1-a_i^t
\right)
b_i^t.
\end{equation}
An active node updates its forwarding budget using the newly received
information mass, whereas a halted node retains its previous budget.
Stronger received information sustains later propagation, while weak
information mass reduces subsequent forwarding capacity.

\paragraph{Network parameterization.}
The observation predictor $f_{\mathrm{pred}}$, carrier updater
$f_c$, gain predictor $f_{\mathrm{gain}}$, gain-aware halting
controller $f_{\mathrm{halt}}$ and reliability gate
$f_{\mathrm{rel}}$ are implemented as lightweight two-layer MLPs
with PReLU activations and dropout. Layer normalization is applied
to representation-valued outputs, whereas sigmoid functions are
used for scalar or feature-wise gates. The source-reception and
channel-selection policies combine source--target query--key
compatibility scores with lightweight scalar MLPs. Within each
interaction round, these modules share parameters across all nodes,
edges, graph domains and rollout steps, whereas successive
interaction rounds use separate parameter sets.

\subsubsection{Information-Flow Regularization}

The node-level information-flow policy contains three coupled
decisions: which sources to receive from, how much effective
information mass to transmit and how long each node should continue
its rollout. Without additional constraints, the policy may degenerate
into uniformly receiving all neighboring information, transmitting
excessive information mass, or keeping most nodes active until the
maximum horizon. We therefore regularize these three aspects
separately.

\paragraph{Source-reception regularization.}
The first regularizer controls the average source-reception strength:
\begin{equation}
\mathcal L_{\gamma}
=
\frac{1}{T|\mathcal E|}
\sum_{t=0}^{T-1}
\sum_{(u,i)\in\mathcal E}
\gamma_{u\rightarrow i}^{t}.
\end{equation}
This term discourages the policy from assigning uniformly high
reception scores to all neighboring sources. A smaller
$\mathcal L_{\gamma}$ encourages the model to be selective about
which neighbors contribute contextual information rather than
reverting to indiscriminate neighbor aggregation.

\paragraph{Propagation-mass regularization.}
The second regularizer controls the effective information mass
transmitted along graph edges. Recall that
\begin{equation}
\mu_{u\rightarrow i}^{t}
=
a_i^t b_u^t\gamma_{u\rightarrow i}^{t},
\end{equation}
where $a_i^t$ indicates whether the target node is active,
$b_u^t$ is the source forwarding budget and
$\gamma_{u\rightarrow i}^{t}$ is the source-reception strength. We
regularize the average transmitted mass by
\begin{equation}
\mathcal L_{\mu}
=
\frac{1}{T|\mathcal E|}
\sum_{t=0}^{T-1}
\sum_{(u,i)\in\mathcal E}
\mu_{u\rightarrow i}^{t}.
\end{equation}
Unlike $\mathcal L_{\gamma}$, this term penalizes the realized
information flow after accounting for target activity and source
forwarding budget. It discourages excessive propagation even when
individual reception scores remain moderate.

\paragraph{Continuation regularization.}
The third regularizer controls the node-wise rollout length:
\begin{equation}
\mathcal L_a
=
\frac{1}{T|\mathcal V|}
\sum_{t=0}^{T-1}
\sum_{v_i\in\mathcal V}
a_i^{t+1}.
\end{equation}
Since $a_i^{t+1}$ indicates whether node $v_i$ remains active after
step $t$, minimizing $\mathcal L_a$ discourages unnecessarily long
rollouts. This term prevents most nodes from remaining active until
the maximum horizon while allowing nodes with positive predictive
gain to continue collecting contextual information.

Together, the three regularizers constrain node-level information
flow from complementary perspectives. $\mathcal L_{\gamma}$
encourages selective source reception, $\mathcal L_{\mu}$ limits the
effective information mass transmitted along active paths and
$\mathcal L_a$ promotes adaptive early stopping. In this way, the
learned policy is discouraged from degenerating into uniform
all-neighbor aggregation, excessive message transmission, or fixed
full-horizon propagation.

\section{Full Component Ablation Results}
\label{app:full_ablation}
Table~\ref{tab:full_ablation} reports the complete ablation results on all ten node-level datasets. Removing source reception $\gamma$ reduces the average accuracy from 0.4540 to 0.4388, showing that selective source acquisition helps suppress irrelevant neighboring evidence. Removing signal-channel selection $\rho$ causes a larger average drop to 0.4200, with particularly clear degradation on Texas and Cornell. This confirms the importance of adapting the signal type to local topology. The largest performance decrease occurs when prediction--observation feedback is removed, reducing the average accuracy to 0.3646. The degradation is especially pronounced on Cora and Photo, indicating that feedback is central to assessing contextual reliability and correcting node states. Removing gain-aware stopping also lowers the average accuracy to 0.4282. Although the variant slightly improves results on Wisconsin and Chameleon, it degrades performance on most datasets, particularly PubMed, Computers and Photo.

\end{document}